%% file: main.tex
\PassOptionsToPackage{table}{xcolor}
\documentclass[sigconf,nonacm]{acmart}

\usepackage{makecell}
\usepackage{multirow}
\usepackage{array}
\usepackage{adjustbox}
\usepackage{subcaption}
\usepackage{wrapfig}
\usepackage{tcolorbox}
\usepackage{tabularx}
\usepackage{dashrule}
\usepackage{dsfont}
\usepackage{amsmath}
\usepackage{algpseudocode}
\usepackage{enumitem}
\usepackage{float}
\usepackage[figuresleft]{rotating}
\definecolor{papergreen}{RGB}{0,150,0}
\definecolor{sectionblue}{RGB}{220,230,241}
\definecolor{bestcolor}{RGB}{255,225,225}
\definecolor{secondcolor}{RGB}{225,238,252}

\newcommand{\bestval}[1]{\cellcolor{bestcolor}\hspace{1.5pt}#1\hspace{1.5pt}}
\newcommand{\secondval}[1]{\cellcolor{secondcolor}\hspace{1.5pt}#1\hspace{1.5pt}}

\newcommand{\bestkey}{{%
\setlength{\fboxsep}{0.15pt}%
\colorbox{bestcolor}{\rule{0pt}{0.65em}\hspace{0.65em}}}}
\newcommand{\secondkey}{{%
\setlength{\fboxsep}{0.15pt}%
\colorbox{secondcolor}{\rule{0pt}{0.65em}\hspace{0.65em}}}}

\acmConference[WSDM '27]
  {The 20th ACM International Conference on Web Search and Data Mining}
  {February 15--19, 2027}
  {Hong Kong}\acmYear{2027}\copyrightyear{2027}

\title{Are These Modules Worth Their Cost? A Paradigm-Level Accuracy-Cost Analysis of In-context Learning Text-to-SQL}

\author{Jiayan Lin}
\authornote{All authors contributed equally to this research.}
\affiliation{%
  \institution{Jinan University}
  \city{Guangzhou}
  \country{China}
}
\email{ljiayan@stu2023.jnu.edu.cn}

\author{Yujia Liu}
\authornotemark[1]
\affiliation{%
  \institution{Jinan University}
  \city{Guangzhou}
  \country{China}
}
\email{yujialiu@stu2023.jnu.edu.cn}

\author{Zijin Hong}
\authornotemark[1]
\affiliation{%
  \institution{The Hong Kong Polytechnic University}
  \city{Kowloon}
  \country{Hong Kong}
}
\email{zijin.hong@connect.polyu.hk}

\author{Zheng Yuan}
\affiliation{%
  \institution{The Hong Kong Polytechnic University}
  \city{Kowloon}
  \country{Hong Kong}
}
\email{yzheng.yuan@connect.polyu.hk}

\author{Yilin Xiao}
\affiliation{%
  \institution{The Hong Kong Polytechnic University}
  \city{Kowloon}
  \country{Hong Kong}
}
\email{yilin.xiao@connect.polyu.hk}

\author{Hao Chen}
\affiliation{%
  \institution{City University of Macau}
  \city{Taipa}
  \country{Macau}}
\email{sundaychenhao@gmail.com}

\author{Qinggang Zhang}
\affiliation{%
  \institution{Jilin University}
  \city{Changchun}
  \country{China}
}
\email{qinggangzhang@jlu.edu.cn}

\author{Xiao Huang}
\affiliation{%
  \institution{The Hong Kong Polytechnic University}
  \city{Kowloon}
  \country{Hong Kong}
}
\email{xiao.huang@polyu.edu.hk}

\author{Feiran Huang}
\affiliation{%
  \institution{Beihang University}
  \city{Beijing}
  \country{China}
}
\email{huangfr@buaa.edu.cn}

\newcommand{\cmark}{\textcolor{green!85!black}{\scalebox{1.25}{$\checkmark$}}}
\newcommand{\na}{\textcolor{gray}{--}}

\newcolumntype{C}[1]{>{\centering\arraybackslash}m{#1}}
\newcolumntype{L}[1]{>{\raggedright\arraybackslash}m{#1}}

\newtcolorbox{paradigmbox}{
  colback=gray!3, colframe=gray!40,
  boxrule=0.4pt, arc=2pt,
  left=6pt, right=6pt, top=6pt, bottom=6pt
}

\newcommand{\modulesep}{%
  \medskip
  \noindent\textcolor{gray!55}{\hdashrule[0pt]{\linewidth}{0.4pt}{2.5pt}}%
  \par\medskip
}

\newcommand{\modulepara}[1]{%
  \par\vspace{1.2ex}%
  \noindent\textbf{#1}%
}

\newif\ifincludeappendix
\includeappendixtrue

\begin{document}

\input{tex/abstract}
\ccsdesc[500]{Information systems~Relational database query languages}
\ccsdesc[300]{Computing methodologies~Natural language processing}
\keywords{Text-to-SQL, In-context Learning, Large Language Models, Accuracy-Cost Trade-offs}
\maketitle

\input{tex/introduction}

\input{tex/taxonomy}
\input{tex/setup}

\input{tex/marginal}

\input{tex/interaction}

\input{tex/guideline}

\input{tex/related}

\input{tex/conclusion}

\input{tex/ethical_considerations}

\nocite{*}
\bibliographystyle{ACM-Reference-Format}
\bibliography{reference}

\ifincludeappendix
  \clearpage
  \appendix
  \input{tex/appendix/full_taxonomy}
  \input{tex/appendix/paradigm_implementation_details}
  \input{tex/appendix/pricing}
  \input{tex/appendix/main_results}
\fi

\end{document}

%% file: tex/abstract.tex
\begin{abstract}
Recent advances in in-context learning (ICL) text-to-SQL have substantially improved execution accuracy on public benchmarks by assembling increasingly elaborate pipelines around the base generator, yet existing studies typically report aggregate end-to-end accuracy, \textbf{without quantifying the marginal accuracy-cost contribution of individual design choices}. Consequently, providing a unified, paradigm-level cost-accuracy quantification remains a critical challenge for understanding and configuring modern text-to-SQL. To address this, we instantiate 17 paradigm-level configurations across five recurring modules of the ICL text-to-SQL pipeline under a single controlled implementation, and attribute each paradigm's marginal contribution and incurred cost across all four backbones spanning diverse capability levels and reasoning styles. Our analysis reveals that execution-feedback refinement is the only paradigm whose benefit holds universally at consistently low cost, while most other modules help only under backbone-dependent conditions. Token accounting shows that input demand is more closely tied to pipeline structure, whereas output demand is more sensitive to backbone generation behavior. Cross-module analysis further shows that stacking improves accuracy on most backbones, although how the gains compose varies with backbone capability. We also find that \textbf{a fixed budget is often better spent engineering a more elaborate pipeline over a mid-tier backbone than upgrading to a frontier model with a lean pipeline}. These findings distill into an actionable, cost-aware tiered guideline that transfers to five additional backbones without per-paradigm search.
\end{abstract}

%% file: tex/introduction.tex
\section{Introduction}

The advent of large language models (LLMs) has driven remarkable progress on text-to-SQL~\cite{hong2025next}, a foundational task that translates natural language questions into executable SQL queries over relational databases. In particular, in-context learning (ICL) approaches~\cite{wang2025agentarscalesql} have substantially pushed execution accuracy on widely used benchmarks such as BIRD~\cite{li2023BIRD} and Spider~\cite{yu2018Spider} by assembling increasingly elaborate pipelines around the base generator. ICL is particularly suitable for such analysis because its improvements are largely realized through inference-time pipeline design, allowing the marginal benefit and added cost of individual components to be measured while holding the underlying backbone fixed. A modern ICL text-to-SQL system rarely issues a single prompt: it retrieves in-context demonstrations~\cite{shi2025gensql}, links the relevant schema~\cite{li2025alphasql}, samples and selects among multiple candidates~\cite{lee2025mcs}, and refines the chosen query through execution feedback or multi-agent coordination~\cite{wang2025mac}. Each successive system stacks more such modules atop the last, propelling state-of-the-art accuracy to new heights on public leaderboards. These increasingly modular pipelines still create a growing design space in which accuracy gains may often arise in practice from very different combinations of inference-time components.

\begin{figure}[!t]
  \centering
  \includegraphics[width=\columnwidth]
    {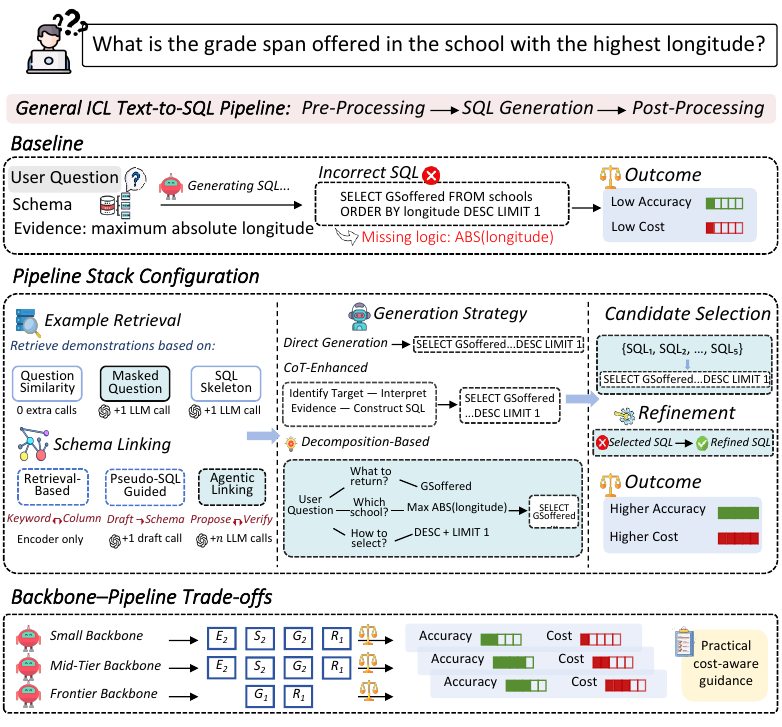}
  \vspace{-4mm}
  \caption{Overview of a modular ICL Text-to-SQL pipeline and its accuracy--cost trade-offs. Five recurring modules span
  pre-processing, SQL generation, and post-processing, but their benefits and inference costs vary across backbones, yielding
  different cost-efficient configurations under fixed budgets.}
  \label{fig:introduction}
  \Description{A schematic comparison of rising Text-to-SQL leaderboard accuracy and increasing token and dollar costs as additional pipeline modules are added.}
  \vspace{-4mm}
\end{figure}

\begin{figure*}[!t]
  \centering
  \includegraphics[width=\linewidth]{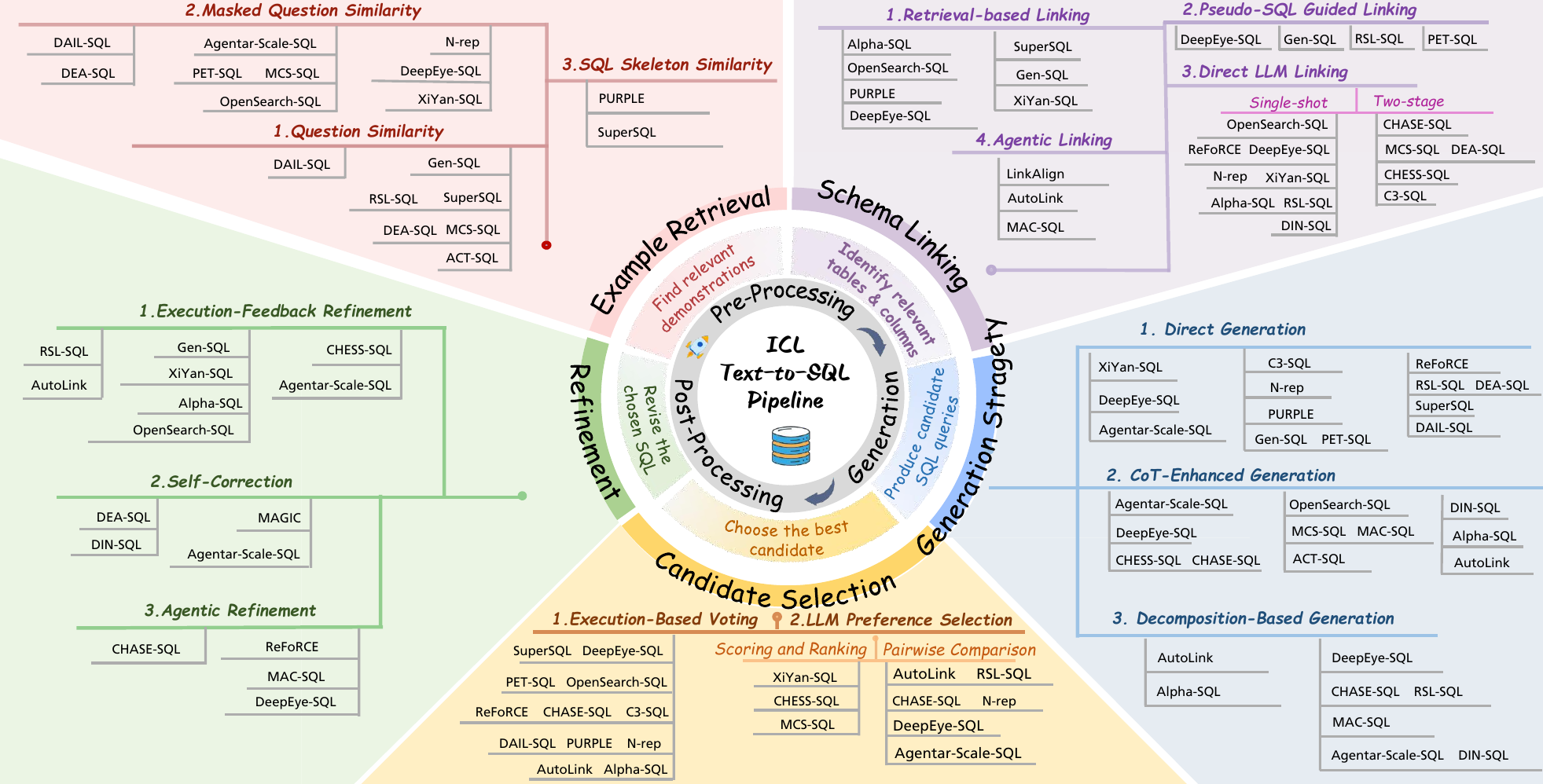}
  \vspace{-3mm}
  \caption{Taxonomy of ICL Text-to-SQL pipelines, organizing five modules across three pipeline stages into 17 paradigm-level configurations and 24 representative methods, including methods that naturally span multiple paradigms within the pipeline. Repeated method names indicate systems that combine design paradigms across modules or use multiple strategies within the same module, providing a unified view of how modern ICL Text-to-SQL systems assemble pipeline components in practice.}
  \label{fig:taxonomy}
  \Description{A circular taxonomy diagram organizing five ICL text-to-SQL modules across preprocessing, SQL generation, and post-processing stages, with branches for 17 paradigm-level configurations and representative methods.}
  \vspace{-3mm}
\end{figure*}

However, \textbf{is the accuracy growth on text-to-SQL leaderboards reliably worth its disproportionately growing cost?} Figure~\ref{fig:introduction} illustrates a concerning trend: execution accuracy keeps rising as more modules are stacked, while the cost required to gain each additional point grows disproportionately. Pipeline components, however, are typically reported only through a single end-to-end accuracy number for the system. This aggregate-only reporting practice raises two concerns:
\textbf{1) The marginal contribution of individual modules is overshadowed by aggregate accuracy}, as the same end-to-end number may be reached through entirely different module combinations, leaving unclear which paradigms carry the gain and which are merely along for the ride; \textbf{2) Cost is rarely accounted for jointly with accuracy}, despite agentic refinement loops and multi-candidate sampling consuming several times, and in some cases an order of magnitude, more tokens and dollars than a single-pass baseline. These concerns make it difficult for practitioners to decide which modules are worth adding for a backbone, and how far the pipeline should go before engineering ceases to pay off. They therefore motivate us to study this recurring class of modular ICL pipelines under controlled settings, where inference-time modules can be varied while holding the backbone fixed to more reliably attribute their marginal accuracy and cost.

Studies have begun to interrogate aspects of text-to-SQL pipelines, such as ablations on schema linking~\cite{maamari2024death} or comparisons among candidate selection strategies~\cite{pourreza2025chase}, while several surveys catalog representative methods and their design choices~\cite{li2024dawn}. However, these efforts either evaluate isolated components within a single system or report each method as a monolith without disentangling its constituent paradigms, leaving open how each module's contribution shifts across backbones of differing capability and reasoning style. In this context, \textbf{effectively quantifying the marginal accuracy--cost tradeoff of each design paradigm in ICL text-to-SQL pipelines remains a significant challenge}. As a solution, this paper conducts a controlled paradigm-level evaluation that decomposes the ICL text-to-SQL pipeline into recurring modules, instantiates each design choice as a paradigm-level configuration under a unified implementation, and isolates its marginal contribution alongside the incurred cost across backbones spanning diverse capability levels and reasoning styles. Unlike accuracy-only evaluations, we jointly examine execution accuracy and inference cost, revealing how each paradigm's value shifts with model capability. Overall, our contributions in this paper are listed as follows:
\begin{itemize}[
  leftmargin=2.0em,
  topsep=2pt,
  partopsep=0pt,
  itemsep=2pt,
  parsep=0pt
]
\item We conduct the first controlled, paradigm-level accuracy--cost evaluation of in-context learning (ICL) text-to-SQL pipelines, instantiating 17 paradigm-level configurations across five recurring modules under a single unified implementation.
\item We reveal that execution-feedback refinement is the only paradigm whose benefit holds universally at consistently low cost, while most other paradigms instead help only under backbone-dependent conditions, exposing the unreliability of aggregate-only accuracy reporting in practice.
\item We further show that a fixed budget is often better spent engineering a more elaborate pipeline over a mid-tier backbone than upgrading to a frontier model with a lean pipeline, and distill these findings into a tiered cost-efficient configuration guideline validated on five additional backbones across capability levels without per-paradigm search.
\end{itemize}

%% file: tex/taxonomy.tex
\section{Pipeline Decomposition}
\label{sec:pipeline}

We decompose the ICL Text-to-SQL pipeline into three main stages, \emph{Pre-Processing}, \emph{SQL Generation}, and \emph{Post-Processing}, comprising five recurring functional modules: \textbf{Example Retrieval}, \textbf{Schema Linking}, \textbf{Generation Strategy}, \textbf{Candidate Selection}, and \textbf{Refinement}. These yield 17 paradigm-level configurations; Figure~\ref{fig:taxonomy} summarizes the taxonomy and 24 representative methods. For a question $q_i$ with schema $\mathcal{S}_i$, the five modules are formalized below.

\modulepara{(1) Example Retrieval.} Retrieves a demonstration set $\mathcal{Z}_i$ from $\mathcal{D}_{\mathrm{train}}$ using a paradigm-specific representation $\rho_e$:
\begin{equation}
    \mathcal{Z}_i
    =
    \operatorname*{TopK}_{z\in\mathcal{D}_{\mathrm{train}}}
    \operatorname{sim}
    \bigl(\rho_e(q_i),\rho_e(q_z)\bigr).
    \label{eq:module-retrieval}
\end{equation}
Here, $q_z$ denotes the question in example $z$.
\emph{$E_1$ Question Similarity} directly uses raw question embeddings~\cite{li2024dawn};
\emph{$E_2$ Masked Question Similarity} masks schema-specific entities before retrieval~\cite{gao2023dailsql}; and
\emph{$E_3$ SQL Skeleton Similarity} uses structural SQL query forms~\cite{ren2024purple}.

\modulepara{(2) Schema Linking.}
For each schema element $u\in\mathcal{S}_i$, a paradigm-specific rule $\ell_s(u\mid q_i,\mathcal{S}_i)\in\{0,1\}$ determines whether $u$ is retained:
\begin{equation}
    \widetilde{\mathcal{S}}_i
    =
    \left\{
        u\in\mathcal{S}_i
        \;\middle|\;
        \ell_s(u\mid q_i,\mathcal{S}_i)=1
    \right\}
    \subseteq \mathcal{S}_i.
    \label{eq:module-linking}
\end{equation}
\emph{$S_1$ Retrieval-Based Linking} scores relevant schema elements via dense retrieval~\cite{xie2025open};
\emph{$S_2$ Pseudo-SQL Guided Linking} retains elements explicitly referenced by a preliminary SQL draft~\cite{shi2025gensql};
\emph{$S_3$ Direct LLM Linking} uses \emph{$S_{3a}$ Single-shot}~\cite{pourreza2023din}
or \emph{$S_{3b}$ Two-stage}~\cite{dong2023c3} LLM linking;
and \emph{$S_4$ Agentic Linking} uses iterative proposer-verifier exchanges.

\modulepara{(3) Generation Strategy.} Generates a pool $\mathcal{Y}_i$ of $N$ SQL candidates:
\begin{equation}
    \mathcal{Y}_i
    =
    \left\{\hat{y}_i^{(k)}\right\}_{k=1}^{N}
    =
    \operatorname{Generate}_g\!\left(
        q_i,\widetilde{\mathcal{S}}_i,\mathcal{Z}_i;N
    \right).
    \label{eq:module-generation}
\end{equation}
Disabled retrieval uses $\mathcal{Z}_i=\varnothing$, while disabled
linking uses $\widetilde{\mathcal{S}}_i=\mathcal{S}_i$.
\emph{$G_1$ Direct Generation} generates SQL directly~\cite{deng2025reforce};
\emph{$G_2$ CoT-Enhanced Generation} elicits intermediate reasoning~\cite{pourreza2025chase};
and \emph{$G_3$ Decomposition-Based Generation} decomposes the question before generating the final SQL~\cite{wang2025agentarscalesql}. They may also be combined via parallel generators.

\modulepara{(4) Candidate Selection.} Selects one query from $\mathcal{Y}_i$ according to selection paradigm $c$:
\begin{equation}
    \hat{y}_i^{\mathrm{sel}}
    =
    \operatorname{Select}_{c}\!\left(
        \{\hat{y}_i^{(k)}\}_{k=1}^{N}
    \right)
    \in \mathcal{Y}_i .
    \label{eq:module-selection}
\end{equation}
\emph{$C_1$ Execution-Based Voting} selects from the largest result-equivalence group~\cite{li2025pet};
\emph{$C_2$ LLM-Based Preference Selection} uses an LLM to choose via \emph{$C_{2a}$ Scoring and Ranking}~\cite{gao2024xiyan} or
\emph{$C_{2b}$ Pairwise Comparison}~\cite{li2026deepeye}.

\modulepara{(5) Refinement.} Maps the selected SQL $\hat{y}_i^{\mathrm{sel}}$ to the final prediction under refinement paradigm $r$:
\begin{equation}
    \hat{y}_i
    =
    \operatorname{Refine}_r\!\left(
        \hat{y}_i^{\mathrm{sel}},
        q_i,
        \widetilde{\mathcal{S}}_i
    \right).
    \label{eq:module-refinement}
\end{equation}
\emph{$R_1$ Execution-Feedback Refinement} rewrites only after an execution error or empty result~\cite{cao2024rsl}; \emph{$R_2$ Self-Correction} performs prompt-level review without execution~\cite{xie2024decomposition}; and
\emph{$R_3$ Agentic Refinement} coordinates multiple revision roles using execution feedback~\cite{wang2025mac}.
Without refinement, the selected SQL is simply returned unchanged~\cite{hong2026errorllm,qu2025share}.

Our taxonomy covers 24 representative ICL Text-to-SQL methods published between 2023 and 2026 that report on BIRD or Spider and provide sufficient pipeline detail for reliable paradigm-level classification. Methods may span multiple paradigms, especially through parallel generation or combined schema linking signals.

%% file: tex/setup.tex
\section{Experimental Setup}
\label{sec:setup}

\noindent
This section describes the experimental setup for our paradigm-level analysis. We first introduce the datasets and backbone scope, then present the evaluation metrics and statistical testing, and finally detail the controlled implementation of the 17 paradigm configurations to ensure consistent and comparable assessment across paradigms and backbones within our unified framework.

\subsection{Dataset and Backbones}
\label{sec:setup:data}

\noindent\textbf{Dataset.} We conduct the main paradigm-level analysis on the development split of BIRD~\cite{li2023BIRD}, which contains 1{,}534 question and SQL pairs over 11 databases. BIRD is widely used in recent ICL Text-to-SQL work and provides evidence annotations that are directly relevant to several paradigms studied in this paper. To examine cross-benchmark generalization, we additionally use Spider~\cite{yu2018Spider} (1{,}034 samples) to evaluate cross-benchmark transfer of the BIRD-derived pipeline configurations under the same protocol in \S~\ref{sec:tradeoffs:stack}.

\modulepara{Backbones.} We run the full set of paradigm configurations on four primary backbones spanning different capability levels and reasoning behaviors: GPT-4o-mini, Gemini-2.5-Flash, DeepSeek-V4-Flash (reasoning), and GPT-5.4. To further examine the effect of reasoning behavior on generation strategies, we additionally evaluate Qwen3-235B in thinking mode and o3-mini in \S~\ref{sec:marginal:dependent}. Finally, we evaluate the tiered guideline in \S~\ref{sec:guideline} on five additional backbones by applying the three prescribed tiers without per-paradigm search.

\subsection{Metrics and Statistical Testing}
\label{sec:setup:metrics}

\noindent\textbf{Accuracy metrics.} For intuitive comparison, we follow the standard evaluation setting in the text-to-SQL community~\cite{yu2018Spider}. Our primary evaluation metric is \textbf{execution accuracy ($\mathrm{EX}$)}. Given a set $\mathcal{D} = \{(q_i, y_i)\}_{i=1}^{n}$ of question and gold-SQL pairs and predicted SQLs $\{\hat{y}_i\}_{i=1}^{n}$:
\begin{equation}
\mathrm{EX} \;=\; \frac{1}{n} \sum_{i=1}^{n} \mathds{1}\!\left[\,\operatorname{\textsc{Exec}}(\hat{y}_i) = \operatorname{\textsc{Exec}}(y_i)\,\right],
\end{equation}
where $\operatorname{\textsc{Exec}}(\cdot)$ denotes the execution operator on the corresponding database. For configurations that sample multiple candidates, let $\{\hat{y}_i^{(k)}\}_{k=1}^{N}$ denote the $N$ sampled candidates for example $i$. We additionally report \textbf{oracle candidate recall ($\mathrm{OCR}$)}:
\begin{equation}
\mathrm{OCR} \;=\; \frac{1}{n} \sum_{i=1}^{n} \bigvee_{k=1}^{N} \mathds{1}\!\left[\,\operatorname{\textsc{Exec}}(\hat{y}_i^{(k)}) = \operatorname{\textsc{Exec}}(y_i)\,\right],
\end{equation}
which upper-bounds the $\mathrm{EX}$ achievable by any selection strategy over a fixed candidate pool.

\modulepara{Cost and resource metrics.} To compare paradigms along the accuracy--cost tradeoff, for a paradigm $p$ instantiated in module $m$, we report \textbf{cost per $\mathrm{EX}$ point ($\mathrm{CPP}$)}:
\begin{equation}
\mathrm{CPP}(p)
=
\frac{\mathrm{Cost}(p)-\mathrm{Cost}(b_m)}
{100\left[\mathrm{EX}(p)-\mathrm{EX}(b_m)\right]},
\end{equation}
where $b_m$ denotes the matched reference for module $m$, using the same sampling budget as $p$ ($N{=}5$ for Candidate Selection and $N{=}1$ otherwise). Measured in USD/pp, $\mathrm{CPP}$ represents the additional cost to gain one percentage point of $\mathrm{EX}$ over the matched reference. $\mathrm{CPP}$ is reported only when $p$ strictly improves $\mathrm{EX}$ at strictly higher cost; paradigms that Pareto-dominate the reference are marked separately. We report all costs in USD and also record mean per-question input (IN) and output (OUT) token counts for comparison to better characterize resource overhead beyond aggregate API cost.

\modulepara{Statistical testing.} All $\mathrm{EX}$ gains over the matched reference are evaluated using McNemar's test. We report 95\% confidence intervals and determine statistical significance at $\alpha=0.05$; detailed numerical intervals are provided in the supplementary material for reference.

\subsection{Controlled Implementation}
\label{sec:setup:impl}

To ensure that quantitative comparisons reflect paradigm-level differences rather than incidental properties of specific systems~\cite{hong2025next}, we re-implement the 17 configurations identified in \S~\ref{sec:pipeline} as paradigm representatives under a unified controlled framework. Rather than reproducing any particular system, each configuration instantiates the defining mechanism of one paradigm while keeping shared implementation settings fixed within corresponding module across all backbones. Figure~\ref{fig:controlled-instantiation} summarizes these controlled implementations across all five modules. Detailed implementation settings and fallback behavior are provided in the final supplementary material.
\modulepara{Baseline and single-swap protocol.} We define a global baseline for each backbone with no retrieved demonstrations, the full schema, direct generation ($G_1$) with $N{=}1$, and neither Candidate Selection nor Refinement. For each target module, we replace only that module with the paradigm under study while keeping other modules at their trivial settings, so that the observed difference can then be attributed directly to the target paradigm. Candidate Selection is the only exception to the $N{=}1$ reference: because selection requires multiple candidates, we use a matched $N{=}5$ baseline that returns the first candidate from the same candidate pool, thereby also separating the selector's contribution from the gain due to multi-sample generation. For Generation Strategy, we additionally evaluate $N{=}5$ sampling to examine effect on candidate-pool quality.

%% file: tex/marginal.tex
\section{Single-Module Analysis}
\label{sec:marginal}
\input{tables/table_controlled_instantiation}
\input{tables/table1}

\begin{figure}[!t]
  \centering
  \includegraphics[width=\columnwidth]{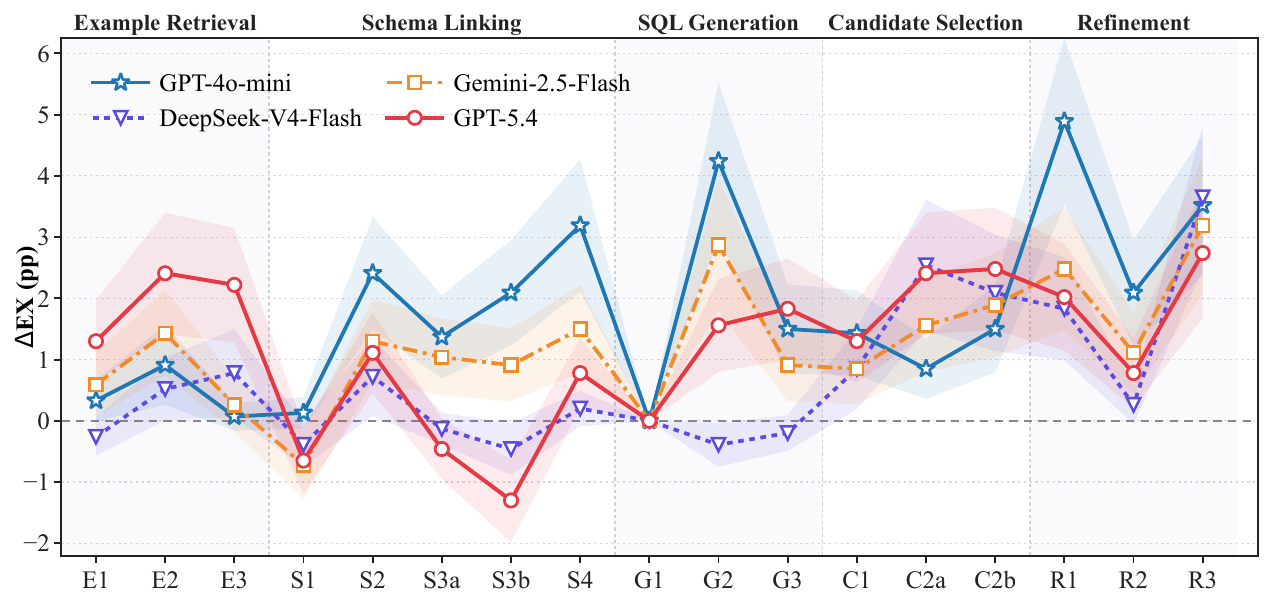}
  \vspace{-5mm}
  \caption{$\Delta\mathrm{EX}$ (pp) relative to the matched-budget reference baseline for 17 paradigm configurations on BIRD dev. Shaded bands show 95\% CIs for paired differences, and the dashed line marks $\Delta\mathrm{EX}=0$. Statistical significance is assessed using McNemar's test. Full numerical CIs are provided in the supplementary material for every paradigm on each backbone.}
  \label{fig:ex-profile}
  \Description{A line chart showing execution-accuracy changes for 17 paradigm configurations across GPT-4o-mini, DeepSeek-V4-Flash, Gemini-2.5-Flash, and GPT-5.4, with 95 percent confidence bands and a zero-change baseline.}
  \vspace{-5mm}
\end{figure}

We instantiate each of the 17 paradigm configurations by replacing exactly one module in the global baseline pipeline, holding the remaining four at their trivial settings. This single-swap protocol isolates the \emph{marginal contribution} of each paradigm. Table~\ref{tab:marginal} reports the accuracy--cost and token statistics across all four backbones, while Figure~\ref{fig:ex-profile} summarizes the $\Delta\mathrm{EX}$ results with 95\% confidence intervals. We organize the analysis around two questions below:

\begin{itemize}
  \item[\textbf{Q1.}] Are there paradigms whose marginal benefit is robust across all four backbones, making them safe pipeline defaults?
  \item[\textbf{Q2.}] For paradigms whose benefit is \emph{not} universal, does the variation correlate with model capability or reasoning style?
\end{itemize}

\subsection{Universal Paradigms}
\label{sec:marginal:universal}

Across the 17 overall paradigm configurations, \textit{\textbf{$R_1$ (Execution-Feedback Refinement) is the only paradigm whose benefit holds across all four backbones at consistently low cost.}} It yields gains between $+1.83$ and $+4.89$~pp, with $\mathrm{CPP}$ below $0.30$~USD/pp throughout. The mechanism is: a diagnostic execution is run on the candidate query, and the model is guided to rewrite only when execution errors or returns no rows. The call is therefore failure-gated rather than mandatory, which keeps the average overhead small while preserving the upside on the queries that actually need revision.

The other two refinement paradigms illustrate why $R_1$'s combination is rare. $R_2$ (Self-Correction) significantly improves $\mathrm{EX}$ on three backbones but has no significant effect on the reasoning backbone DeepSeek-V4-Flash ($+0.26$~pp, n.s.); even where it helps, its $\mathrm{CPP}$ is far above $R_1$'s. $R_3$ (Agentic Refinement) is positive on all four backbones and posts the highest absolute $\mathrm{EX}$ in several columns, but at $\mathrm{CPP}$ one to two orders of magnitude above $R_1$. The pattern across the three paradigms is consistent: execution feedback supplies information the model did not have at generation time, while prompt-internal correction adds only what the model could in principle have produced on its own. Among universally helpful paradigms, this information asymmetry, rather than reasoning quality, is thus what separates cheap from expensive in this setting.
\vspace{-6mm}
\subsection{Model-Dependent Paradigms}
\label{sec:marginal:dependent}

The remaining four modules deliver gains that depend on the backbone. Rather than treating each module in isolation, we organize the discussion by backbone characteristics, since the modules' contributions move together with model capability across tested backbones.

\input{tables/table_reasoning}

\modulepara{Capability gradient: schema linking and example retrieval exhibit opposing trends.} On GPT-4o-mini, the weakest backbone in our set, Schema Linking delivers its largest gains across the four backbones: $S_4$ (Agentic Linking) adds $+3.19$~pp, and $S_2$, $S_{3a}$, $S_{3b}$ all clear $+1.3$~pp. The benefit shrinks sharply on stronger backbones. On GPT-5.4, only $S_2$ and $S_4$ yield positive gains, both at substantially higher $\mathrm{CPP}$ than pre-processing paradigms typically incur on cheaper backbones. \textit{\textbf{Schema linking acts as a capability prosthesis: its value falls as the backbone's schema-grounding ability rises.}}  Recent studies~\cite{maamari2024death,yuan2025knapsack} report that sufficiently strong models can generate correct SQL even without explicit schema linking; we extend the observation to a four-backbone, cost-aware setting. Example Retrieval shows the opposite trend, though weakly. $E_2$ (Masked Question Similarity) is the best retrieval paradigm on every non-reasoning backbone, and its gain scales with baseline $\mathrm{EX}$, from $+0.91$~pp on GPT-4o-mini to $+2.41$~pp on GPT-5.4. One reading is that stronger non-reasoning backbones extract reusable structure from demonstrations, but DeepSeek-V4-Flash departs from this trend in ways we cannot attribute to capability alone in this case. These trends suggest backbone-dependent preprocessing benefits.

\modulepara{Reasoning split: external reasoning scaffolds help mainly non-reasoning backbones.} Among the four primary backbones tested, Generation Strategy varies more than any other module. $G_2$ (CoT-Enhanced Generation) delivers $+4.24$~pp on GPT-4o-mini at $\mathrm{CPP}=0.026$~USD/pp, the second-largest single-module gain after $R_1$. The gain shrinks on Gemini-2.5-Flash and GPT-5.4, and turns negative on DeepSeek-V4-Flash ($-0.39$~pp), with $\mathrm{CPP}$ on reasoning-capable backbones one to two orders of magnitude higher. $G_3$ (Decomposition-Based Generation) follows a similar pattern, with negligible effect on DeepSeek-V4-Flash; the exception is GPT-5.4, where $G_3$ slightly exceeds $G_2$ ($+1.83$ vs.\ $+1.56$~pp) at higher $\mathrm{CPP}$.

\begin{figure}[!t]
  \centering
  \includegraphics[width=\columnwidth]{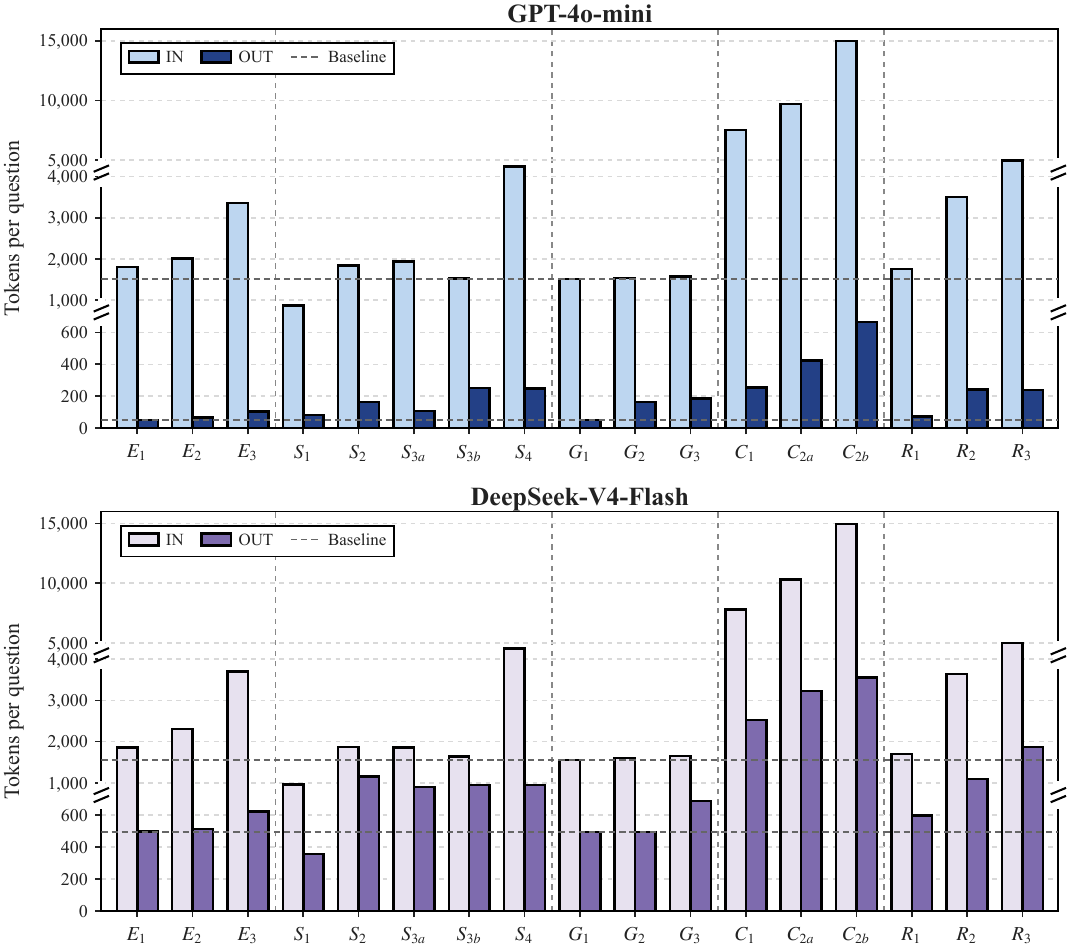}
  \vspace{-6mm}
  \caption{Per-question input (IN) and output (OUT) token consumption across the 17 paradigm configurations for two backbones, GPT-4o-mini and DeepSeek-V4-Flash, on BIRD. Dashed horizontal lines mark the corresponding baseline token levels for direct comparison across configurations.}
  \label{fig:tokenbars}
  \vspace{-6mm}
\end{figure}

\input{tables/table_pairs}

Because DeepSeek-V4-Flash is the only reasoning backbone among the primary four models, we further compare all three generation strategies on Qwen3-235B in thinking mode and o3-mini (Table~\ref{tab:reasoning_generation}). On Qwen3-235B, $G_2$ changes $\mathrm{EX}$ by $+0.20$~pp and $G_3$ by $-0.52$~pp. On o3-mini, the changes are $-0.65$ and $-1.04$~pp, with higher cost in every case. \textit{\textbf{Across the three reasoning backbones tested, external reasoning scaffolds provide no consistent accuracy gain over direct generation, despite higher cost.}} We nevertheless treat this split as suggestive rather than universal, since reasoning style, capability, and model family remain confounded.

\modulepara{Pool-bound regime: candidate selection saturates regardless of backbone.} All three selectors significantly improve $\mathrm{EX}$ over the $N{=}5$ baseline on every backbone, but we do not test selector against selector; we therefore draw no ranking among $C_1$, $C_{2a}$, $C_{2b}$. More consequentially, taking the best selector per backbone, a substantial gap to the oracle ceiling remains everywhere, reaching $5.74$~pp on GPT-4o-mini. \textit{\textbf{Candidate selection is recall-bound: once the candidate pool is fixed, further gains must come from improving the pool itself or revising the chosen query post-hoc.}} The fixed-pool recall ceiling, together with the substantial combined inference cost of multi-candidate generation and selection, motivates the exclusion of Candidate Selection from both the stacking analysis in \S~\ref{sec:tradeoffs} and the cost-aware configuration guideline in \S~\ref{sec:guideline}.

\subsection{Token-Level Resource Decomposition}
\label{sec:token-decomposition}

USD cost alone does not reveal where a paradigm's resource overhead comes from. Table~\ref{tab:marginal} reports the complete IN/OUT token accounting across all four backbones, while Figure~\ref{fig:tokenbars} visualizes two contrasting backbones. The source of token overhead closely follows the paradigm mechanism. Example Retrieval and LLM-based Candidate Selection are primarily input-heavy: $E_3$ substantially expands the prompt context, while $C_{2b}$ reaches about 15K input tokens. In contrast, Generation Strategy mainly shifts resource use toward output consumption through additional reasoning or decomposition. Schema Linking is more heterogeneous: $S_1$ reduces input through schema pruning, whereas $S_4$ incurs substantially larger input context from iterative interaction. Refinement shows a different pattern: $R_1$ remains close to the baseline in both IN and OUT, whereas $R_2$ and $R_3$ incur substantially larger token overhead.

Across backbones, token consumption varies more strongly in OUT than IN. For a given paradigm, IN is stable across the four backbones, while the two-backbone visualization shows a pronounced OUT gap: DeepSeek-V4-Flash consistently produces substantially more OUT than GPT-4o-mini, with its reasoning traces contributing to the OUT total. This pattern suggests that input demand is more closely tied to pipeline structure, whereas output demand is more sensitive to backbone generation behavior, especially reasoning.

\subsection{Single-Module Conclusion}
\label{sec:marginal:takeaways}

\textit{\textbf{The single-module analysis surfaces one safe default and a structured account of when the rest help and where their resource overhead arises.}} Only $R_1$ qualifies as a universal, low-cost default; other modules' contributions are backbone-dependent. The dependence has structure: input-side scaffolding tracks capability, generation-side scaffolding tracks reasoning style, and post-hoc selection saturates against a pool-bound ceiling. Resource use shows a complementary pattern: IN is more closely tied to pipeline structure, whereas OUT is more sensitive to backbone generation behavior. \S~\ref{sec:tradeoffs} next asks whether these effects compose when paradigms are stacked, and how a fixed budget should be allocated across stages.

%% file: tables/table_controlled_instantiation.tex
\begin{figure}[t]
\centering

\begingroup
\footnotesize

\noindent\rule{\linewidth}{0.6pt}\par

\noindent\parbox[c][2.0em][c]{\linewidth}{%
  \centering
  \small\bfseries Controlled Paradigm Implementation%
}\par

\noindent\rule{\linewidth}{0.6pt}\par
\vspace{2pt}

\begin{algorithmic}[1]

\Require $q_i$, $\mathcal{S}_i$, optional evidence, backbone,
target module $m$, and paradigm setting
\Ensure final prediction $\hat{y}_i$

\State Keep all non-target modules at their trivial settings.

\If{$m$ is \textsc{Example Retrieval}}
    \State \textcolor{black!55}{\textit{Fix $k{=}3$, the training retrieval pool, and the same frozen encoder.}}
    \State
    $\displaystyle
    \mathcal{Z}_i \gets
    \begin{cases}
        \text{retrieve with } q_i,
            & E_1,\\
        \text{mask schema entities} \rightarrow \text{retrieve},
            & E_2,\\
        \text{draft SQL} \rightarrow \text{skeleton} \rightarrow \text{retrieve},
            & E_3.
    \end{cases}$

\ElsIf{$m$ is \textsc{Schema Linking}}
    \State \textcolor{black!55}{\textit{Fix the full-schema input and always retain primary/foreign keys.}}
    \State
    $\displaystyle
    \widetilde{\mathcal{S}}_i \gets
    \begin{cases}
        \text{retrieval-based filtering},
            & S_1,\\
        \text{draft SQL} \rightarrow \text{referenced schema},
            & S_2,\\
        \text{one-shot table/column selection},
            & S_{3a},\\
        \text{tables} \rightarrow \text{columns},
            & S_{3b},\\
        \text{propose} \rightarrow \text{verify} \rightarrow [\text{revise}],
            & S_4.
    \end{cases}$

\ElsIf{$m$ is \textsc{Generation Strategy}}
    \State \textcolor{black!55}{\textit{Fix one backbone generation call at temperature $0$.}}
    \State
    $\displaystyle
    \mathcal{Y}_i \gets
    \begin{cases}
        \text{generate SQL directly},
            & G_1,\\
        \text{reason} \rightarrow \text{generate final SQL},
            & G_2,\\
        \text{decompose} \rightarrow \text{generate final SQL},
            & G_3.
    \end{cases}$

\ElsIf{$m$ is \textsc{Candidate Selection}}
    \State \textcolor{black!55}{\textit{Fix the same $N{=}5$ pool $\mathcal{Y}_i$ sampled from $G_1$.}}
    \State Execute all candidates in $\mathcal{Y}_i$.
    \State
    $\displaystyle
    \hat{y}_i^{\mathrm{sel}} \gets
    \begin{cases}
        \text{result-equivalence voting},
            & C_1,\\
        \text{score/rank with execution outcomes},
            & C_{2a},\\
        \text{pairwise tournament},
            & C_{2b}.
    \end{cases}$

\ElsIf{$m$ is \textsc{Refinement}}
    \State \textcolor{black!55}{\textit{Fix the same inputs and temperature $0$; keep the last valid SQL on failure.}}
    \State
    $\displaystyle
    \hat{y}_i \gets
    \begin{cases}
        \text{execute} \rightarrow [\text{rewrite if error/empty}],
            & R_1,\\
        \text{review without execution feedback},
            & R_2,\\
        \text{correct with execution feedback}
        \rightarrow \text{verify} \rightarrow [\text{correct}],
            & R_3.
    \end{cases}$
\EndIf

\State Run the remaining pipeline stages at their trivial settings.
\State \Return $\hat{y}_i$

\end{algorithmic}

\vspace{1pt}

\noindent{\scriptsize
\textcolor{black!55}{\textit{Gray italics: shared fixed settings.}}
\quad
$[\cdot]$ denotes a conditional operation.
}\par

\vspace{1pt}
\noindent\rule{\linewidth}{0.6pt}

\endgroup
\vspace{-2mm}
\caption{Controlled implementation of the 17 paradigm configurations. For each target module, shared settings are fixed and all non-target modules remain at their trivial settings, so that only the paradigm-specific mechanism itself varies. }
\label{fig:controlled-instantiation}
\vspace{-4mm}
\end{figure}

%% file: tables/table1.tex
\begin{table*}[t]
\caption{Module-level accuracy--cost results on BIRD dev across four LLMs. \textbf{EX}: execution accuracy (\%); \textbf{Cost}: total API spend (USD); \textbf{CPP}: incremental USD per EX point over the matched-budget baseline ($N{=}5$ for Candidate Selection, $N{=}1$ otherwise; \S~\ref{sec:setup:metrics}). \textbf{IN/OUT}: mean per-question tokens (DeepSeek-V4-Flash reasoning tokens count as OUT). Significance uses McNemar's test ($\alpha{=}0.05$; full CIs in the supplement). In CPP, numbers denote significant gains; ``$^{\circ}$'', nonsignificant gains; ``\textendash'', no EX improvement; and ``$^{*}$'', a significant gain at no higher cost.
\bestkey~/~\secondkey~mark the best/second-best results;
\textbf{OCR@5} is the oracle upper bound for $N{=}5$.}
\label{tab:marginal}
\centering
\scriptsize
\setlength{\tabcolsep}{0.5pt}
\renewcommand{\arraystretch}{1.10}

\begin{tabular*}{\textwidth}{
@{\hspace{8pt}\extracolsep{\fill}}
c
ccccc
ccccc
ccccc
ccccc
@{\hspace{8pt}}
}
\toprule
\multirow{2}{*}{\footnotesize\bfseries Paradigm}
  & \multicolumn{5}{c}{\footnotesize\bfseries GPT-4o-mini}
  & \multicolumn{5}{c}{\footnotesize\bfseries DeepSeek-V4-Flash}
  & \multicolumn{5}{c}{\footnotesize\bfseries Gemini-2.5-Flash}
  & \multicolumn{5}{c}{\footnotesize\bfseries GPT-5.4} \\
\cmidrule(lr){2-6}
\cmidrule(lr){7-11}
\cmidrule(lr){12-16}
\cmidrule(lr){17-21}

& {\footnotesize\textbf{EX}\(\boldsymbol{\uparrow}\)}
& {\footnotesize\textbf{Cost}\(\boldsymbol{\downarrow}\)}
& {\footnotesize\textbf{CPP}\(\boldsymbol{\downarrow}\)}
& {\footnotesize\textbf{IN}\(\boldsymbol{\downarrow}\)}
& {\footnotesize\textbf{OUT}\(\boldsymbol{\downarrow}\)}

& {\footnotesize\textbf{EX}\(\boldsymbol{\uparrow}\)}
& {\footnotesize\textbf{Cost}\(\boldsymbol{\downarrow}\)}
& {\footnotesize\textbf{CPP}\(\boldsymbol{\downarrow}\)}
& {\footnotesize\textbf{IN}\(\boldsymbol{\downarrow}\)}
& {\footnotesize\textbf{OUT}\(\boldsymbol{\downarrow}\)}

& {\footnotesize\textbf{EX}\(\boldsymbol{\uparrow}\)}
& {\footnotesize\textbf{Cost}\(\boldsymbol{\downarrow}\)}
& {\footnotesize\textbf{CPP}\(\boldsymbol{\downarrow}\)}
& {\footnotesize\textbf{IN}\(\boldsymbol{\downarrow}\)}
& {\footnotesize\textbf{OUT}\(\boldsymbol{\downarrow}\)}

& {\footnotesize\textbf{EX}\(\boldsymbol{\uparrow}\)}
& {\footnotesize\textbf{Cost}\(\boldsymbol{\downarrow}\)}
& {\footnotesize\textbf{CPP}\(\boldsymbol{\downarrow}\)}
& {\footnotesize\textbf{IN}\(\boldsymbol{\downarrow}\)}
& {\footnotesize\textbf{OUT}\(\boldsymbol{\downarrow}\)} \\

\multicolumn{21}{l}
{\textbf{\textit{Baselines}}} \\
\midrule
\addlinespace[2pt]

$N{=}1$
& 49.74 & 0.394 & -- & 1510 & 51
& 56.32 & 0.547 & -- & 1557 & 494
& 59.84 & 1.035 & -- & 1642 & 73
& 61.28 & 7.305 & -- & 1509 & 66 \\

$N{=}5$
& 49.67 & 1.972 & -- & 7550 & 255
& 56.45 & 2.756 & -- & 7787 & 2523
& 60.04 & 5.219 & -- & 8222 & 374
& 61.41 & 35.864 & -- & 7545 & 301 \\

$\mathrm{OCR}@5$
& 56.91 & 1.972 & -- & 7550 & 255
& 60.56 & 2.756 & -- & 7787 & 2523
& 64.54 & 5.219 & -- & 8222 & 374
& 66.82 & 35.864 & -- & 7545 & 301 \\

\multicolumn{21}{l}
{\textbf{\textit{Example Retrieval}}} \\
\midrule
\addlinespace[2pt]

\textbf{$E_1$}
& \secondval{50.07} & \bestval{0.462} & $^{\circ}$ & \bestval{1806} & \bestval{50}
& 56.06 & \bestval{0.613} & -- & \bestval{1858} & \bestval{498}
& \secondval{60.43} & \bestval{1.270} & \bestval{0.401} & \bestval{1842} & \bestval{110}
& 62.58 & \bestval{8.602} & \secondval{0.995} & \bestval{1834} & \bestval{68} \\

\textbf{$E_2$}
& \bestval{50.65} & \secondval{0.525} & \bestval{0.144} & \secondval{2009} & \secondval{68}
& \secondval{56.84} & \secondval{0.717} & \bestval{0.326} & \secondval{2311} & \secondval{512}
& \bestval{61.28} & \secondval{1.632} & \secondval{0.417} & \secondval{2207} & \secondval{161}
& \bestval{63.69} & \secondval{9.613} & \bestval{0.957} & \secondval{2058} & \secondval{75} \\

\textbf{$E_3$}
& 49.80 & 0.867 & $^{\circ}$ & 3357 & 103
& \bestval{57.11} & 1.063 & \secondval{0.660} & 3708 & 622
& 60.10 & 2.421 & $^{\circ}$ & 3358 & 228
& \secondval{63.49} & 14.879 & 3.417 & 3325 & 92 \\

\multicolumn{21}{l}
{\textbf{\textit{Schema Linking}}} \\
\midrule
\addlinespace[2pt]

\textbf{$S_1$}
& 49.87 & \bestval{0.275} & $^{\circ}$ & \bestval{873} & \bestval{81}
& 55.93 & \bestval{0.360} & -- & \bestval{964} & \bestval{357}
& 59.13 & \bestval{1.117} & -- & \bestval{857} & \bestval{188}
& 60.63 & \bestval{5.102} & -- & \bestval{826} & \bestval{84} \\

\textbf{$S_2$}
& \secondval{52.15} & 0.574 & \bestval{0.075} & 1844 & 163
& \bestval{57.04} & 0.899 & \bestval{0.491} & 1871 & 1157
& \secondval{61.15} & 1.784 & \bestval{0.575} & 1855 & 243
& \bestval{62.39} & 10.955 & \bestval{3.293} & 1903 & 159 \\

\textbf{$S_{3a}$}
& 51.11 & \secondval{0.544} & 0.109 & 1936 & \secondval{107}
& 56.19 & 0.782 & -- & 1852 & \secondval{896}
& 60.89 & \secondval{1.659} & \secondval{0.599} & 1924 & \secondval{202}
& 60.82 & \secondval{9.672} & -- & 1944 & \secondval{96} \\

\textbf{$S_{3b}$}
& 51.83 & 0.582 & \secondval{0.090} & \secondval{1528} & 250
& 55.87 & \secondval{0.757} & -- & \secondval{1634} & 944
& 60.76 & 1.808 & 0.847 & \secondval{1577} & 282
& 59.97 & 12.806 & -- & \secondval{1622} & 286 \\

\textbf{$S_4$}
& \bestval{52.93} & 1.260 & 0.271 & 4478 & 249
& \secondval{56.52} & 1.380 & $^{\circ}$ & 4520 & 953
& \bestval{61.34} & 2.870 & 1.224 & 4524 & 205
& \secondval{62.06} & 23.631 & \secondval{20.870} & 4568 & 266 \\

\multicolumn{21}{l}
{\textbf{\textit{SQL Generation}}} \\
\midrule
\addlinespace[2pt]

\textbf{$G_1$}
& 49.74 & \bestval{0.394} & -- & \bestval{1510} & \bestval{51}
& \bestval{56.32} & \bestval{0.547} & -- & \bestval{1557} & \secondval{494}
& 59.84 & \bestval{1.035} & -- & 1642 & \bestval{73}
& 61.28 & \bestval{7.305} & -- & \bestval{1509} & \bestval{66} \\

\textbf{$G_2$}
& \bestval{53.98} & \secondval{0.503} & \bestval{0.026} & \secondval{1534} & \secondval{163}
& 55.93 & \secondval{0.556} & -- & \secondval{1602} & \bestval{493}
& \bestval{62.71} & \secondval{1.522} & \bestval{0.170} & \bestval{1538} & \secondval{212}
& \secondval{62.84} & \secondval{10.000} & \bestval{1.723} & \secondval{1533} & \secondval{179} \\

\textbf{$G_3$}
& \secondval{51.24} & 0.534 & \secondval{0.093} & 1574 & 187
& \secondval{56.13} & 0.651 & -- & 1650 & 689
& \secondval{60.76} & 1.925 & \secondval{0.976} & \secondval{1590} & 311
& \bestval{63.10} & 11.476 & \secondval{2.285} & 1573 & 237 \\

\multicolumn{21}{l}
{\textbf{\textit{Candidate Selection}}} \\
\midrule
\addlinespace[2pt]

\textbf{$C_1$}
& \secondval{51.11} & \bestval{1.972} & \bestval{$^{*}$} & \bestval{7550} & \bestval{255}
& 57.30 & \bestval{2.756} & \bestval{$^{*}$} & \bestval{7787} & \bestval{2523}
& 60.89 & \bestval{5.219} & \bestval{$^{*}$} & \bestval{8222} & \bestval{374}
& 62.71 & \bestval{35.864} & \bestval{$^{*}$} & \bestval{7545} & \bestval{301} \\

\textbf{$C_{2a}$}
& 50.52 & \secondval{2.625} & \secondval{0.771} & \secondval{9718} & \secondval{423}
& \bestval{59.00} & \secondval{3.602} & \secondval{0.333} & \secondval{10319} & \secondval{3227}
& \secondval{61.60} & \secondval{8.045} & \secondval{1.806} & \secondval{10018} & \secondval{896}
& \secondval{63.82} & \secondval{53.348} & \secondval{7.249} & \secondval{9964} & \secondval{658} \\

\textbf{$C_{2b}$}
& \bestval{51.17} & 4.056 & 1.389 & 14962 & 666
& \secondval{58.54} & 4.743 & 0.953 & 14962 & 3562
& \bestval{61.93} & 10.622 & 2.858 & 14846 & 988
& \bestval{63.89} & 75.436 & 15.975 & 14933 & 790 \\

\multicolumn{21}{l}
{\textbf{\textit{Refinement}}} \\
\midrule
\addlinespace[2pt]

\textbf{$R_1$}
& \bestval{54.63} & \bestval{0.471} & \bestval{0.016} & \bestval{1750} & \bestval{74}
& \secondval{58.15} & \bestval{0.620} & \bestval{0.040} & \bestval{1697} & \bestval{596}
& \secondval{62.32} & \bestval{1.099} & \bestval{0.026} & \bestval{1671} & \bestval{86}
& \secondval{63.30} & \bestval{7.905} & \bestval{0.297} & \bestval{1592} & \bestval{78} \\

\textbf{$R_2$}
& 51.83 & \secondval{1.030} & 0.305 & \secondval{3509} & 242
& 56.58 & \secondval{1.253} & $^{\circ}$ & \secondval{3638} & \secondval{1098}
& 60.95 & \secondval{3.211} & 1.964 & \secondval{3513} & \secondval{416}
& 62.06 & \secondval{21.918} & 18.680 & \secondval{3518} & \secondval{366} \\

\textbf{$R_3$}
& \secondval{53.26} & 1.365 & \secondval{0.276} & 4980 & \secondval{238}
& \bestval{59.97} & 1.878 & \secondval{0.365} & 5007 & 1870
& \bestval{63.04} & 4.281 & \secondval{1.016} & 5116 & 502
& \bestval{64.02} & 28.578 & \secondval{7.769} & 5251 & 367 \\
\bottomrule
\end{tabular*}

\end{table*}

%% file: tables/table_reasoning.tex
\begin{table}[!t]
\centering
\caption{Generation strategy results on BIRD dev for two additional reasoning backbones. Qwen3-235B uses thinking mode. Parentheses report $\Delta\mathrm{EX}$ over $G_1$ (pp). Units: Cost in USD, CPP in USD/pp, and IN/OUT in tokens per question.}
\label{tab:reasoning_generation}
\vspace{-2mm}
\setlength{\tabcolsep}{3pt}
\renewcommand{\arraystretch}{1.15}

\resizebox{\columnwidth}{!}{%
\begin{tabular}{ccccccc}
\toprule
\textbf{Backbone}
& \textbf{Paradigm}
& \textbf{EX}$\uparrow$
& \textbf{Cost}$\downarrow$
& \textbf{CPP}$\downarrow$
& \textbf{IN}$\downarrow$
& \textbf{OUT}$\downarrow$ \\
\midrule

\multirow{3}{*}{\shortstack[c]{Qwen3-235B\\(Thinking)}}
& $G_1$
& 57.82
& 4.55
& --
& 1555
& 1134 \\

& $G_2$
& 58.02\enspace\textcolor{green!50!black}{($+0.20$)}
& 5.22
& 3.426
& 1529
& 1327 \\

& $G_3$
& 57.30\enspace\textcolor{red!70!black}{($-0.52$)}
& 5.84
& --
& 1596
& 1495 \\

\midrule

\multirow{3}{*}{o3-mini}
& $G_1$
& 54.11
& 5.67
& --
& 1509
& 463 \\

& $G_2$
& 53.46\enspace\textcolor{red!70!black}{($-0.65$)}
& 5.83
& --
& 1546
& 477 \\

& $G_3$
& 53.06\enspace\textcolor{red!70!black}{($-1.04$)}
& 7.08
& --
& 1538
& 665 \\

\bottomrule
\end{tabular}%
}
\vspace{-4mm}
\end{table}

%% file: tables/table_pairs.tex
\begin{table*}[!t]
\centering
\caption{Cross-module interactions across four backbones. EX is combined execution accuracy (\%), with gray values marking combinations that do not outperform the better of the two individual paradigms. Green and red $\mathcal{I}$ denote super- and sub-additive interactions. The $G_1\times R_j$ rows are omitted because they reduce to $R_j$ and have $\mathcal{I}=0$ by construction. Cost is reported in USD. All interaction values are measured in percentage points relative to the matched-budget reference baseline for each backbone.}
\label{tab:pairs}
\vspace{-3mm}
\setlength{\tabcolsep}{3pt}

\resizebox{\textwidth}{!}{%
\begin{tabular}{c*{4}{ccc}}
\toprule
& \multicolumn{3}{c}{\textbf{GPT-4o-mini}}
& \multicolumn{3}{c}{\textbf{DeepSeek-V4-Flash}}
& \multicolumn{3}{c}{\textbf{Gemini-2.5-Flash}}
& \multicolumn{3}{c}{\textbf{GPT-5.4}} \\
\cmidrule(lr){2-4}
\cmidrule(lr){5-7}
\cmidrule(lr){8-10}
\cmidrule(lr){11-13}
\textbf{Combination}
& \textbf{EX}$\uparrow$ & $\boldsymbol{\mathcal{I}}$ & \textbf{Total Cost}$\downarrow$
& \textbf{EX}$\uparrow$ & $\boldsymbol{\mathcal{I}}$ & \textbf{Total Cost}$\downarrow$
& \textbf{EX}$\uparrow$ & $\boldsymbol{\mathcal{I}}$ & \textbf{Total Cost}$\downarrow$
& \textbf{EX}$\uparrow$ & $\boldsymbol{\mathcal{I}}$ & \textbf{Total Cost}$\downarrow$ \\
\midrule

$G_2\times R_1$
& 56.84
& \textcolor{red!70!black}{$-2.02$}
& 0.612
& \textcolor{gray}{57.82}
& \textcolor{green!50!black}{$+0.07$}
& 0.743
& 65.12
& \textcolor{red!70!black}{$-0.07$}
& 1.971
& 64.60
& \textcolor{red!70!black}{$-0.26$}
& 10.310 \\

$G_2\times R_2$
& 55.87
& \textcolor{red!70!black}{$-0.20$}
& 1.143
& \textcolor{gray}{55.22}
& \textcolor{red!70!black}{$-0.97$}
& 1.457
& 63.62
& \textcolor{red!70!black}{$-0.20$}
& 3.856
& 64.73
& \textcolor{green!50!black}{$+1.11$}
& 24.751 \\

$G_2\times R_3$
& 57.17
& \textcolor{red!70!black}{$-0.33$}
& 1.525
& \textcolor{gray}{59.84}
& \textcolor{green!50!black}{$+0.26$}
& 2.004
& 66.04
& \textcolor{green!50!black}{$+0.13$}
& 4.748
& 65.19
& \textcolor{red!70!black}{$-0.39$}
& 32.632 \\

$G_2\times E_2$
& 54.50
& \textcolor{red!70!black}{$-0.39$}
& 0.685
& \textcolor{gray}{56.39}
& \textcolor{red!70!black}{$-0.07$}
& 0.793
& 64.34
& \textcolor{green!50!black}{$+0.20$}
& 2.217
& 65.32
& \textcolor{green!50!black}{$+0.07$}
& 12.190 \\

\addlinespace[2pt]

$G_3\times R_1$
& 55.74
& \textcolor{red!70!black}{$-0.39$}
& 0.657
& \textcolor{gray}{57.89}
& \textcolor{red!70!black}{$-0.07$}
& 0.733
& 63.43
& \textcolor{green!50!black}{$+0.19$}
& 1.956
& 65.38
& \textcolor{green!50!black}{$+0.26$}
& 12.114 \\

$G_3\times R_2$
& 52.28
& \textcolor{red!70!black}{$-1.05$}
& 1.182
& \textcolor{gray}{55.67}
& \textcolor{red!70!black}{$-0.72$}
& 1.501
& \textcolor{gray}{59.45}
& \textcolor{red!70!black}{$-2.42$}
& 4.523
& 64.47
& \textcolor{green!50!black}{$+0.59$}
& 26.360 \\

$G_3\times R_3$
& \textcolor{gray}{52.93}
& \textcolor{red!70!black}{$-1.83$}
& 1.441
& 60.17
& \textcolor{green!50!black}{$+0.39$}
& 2.067
& 64.60
& \textcolor{green!50!black}{$+0.64$}
& 5.304
& 66.56
& \textcolor{green!50!black}{$+0.72$}
& 34.801 \\

\bottomrule
\end{tabular}%
}
\vspace{-3mm}
\end{table*}

%% file: tex/interaction.tex
\section{Cross-Module and -Model Tradeoffs}
\label{sec:tradeoffs}

The single-swap protocol discussed in \S~\ref{sec:marginal} isolates each paradigm's marginal contribution, but two research questions remain unsolved:

\begin{itemize}
  \item[\textbf{Q3.}] When two paradigms are used together, do their gains compose additively, or does their combination yield smaller or larger gains than expected across different backbones?
  \item[\textbf{Q4.}] Given a fixed budget, is it more cost-efficient to engineer a richer pipeline on a cheaper backbone, or instead to simply upgrade to a stronger backbone running a lean pipeline?
\end{itemize}

\subsection{Module Interactions}
\label{sec:tradeoffs:pairs}

We study Q3 through a systematic analysis of module interactions. Since exhaustive coverage is infeasible, we focus on the full $3\times3$ Generation $\times$ Refinement grid across all backbones. We choose this pair because refinement revises the SQL produced by generation, making their interaction particularly direct. For paradigms $A$ and $B$, we quantify their interaction in percentage points as
\begin{equation}
    \mathcal{I} \;=\; \Delta\mathrm{EX}_{\mathrm{pp}}(A\&B) - \Delta\mathrm{EX}_{\mathrm{pp}}(A) - \Delta\mathrm{EX}_{\mathrm{pp}}(B),
\end{equation}
where positive $\mathcal{I}$ indicates super-additive gains and negative $\mathcal{I}$ indicates sub-additive gains, both relative to the baseline pipeline. Table~\ref{tab:pairs} reports the resulting configurations and interaction terms.

\modulepara{Generation $\times$ Refinement.} Of the 24 nontrivial $G_2/G_3\times R_{1\text{--}3}$ configurations, 17 achieve higher $\mathrm{EX}$ than the better individual paradigm. The pattern is strongest on GPT-5.4, where six combinations improve over both individual paradigms; five of six do so on both GPT-4o-mini and Gemini-2.5-Flash. DeepSeek-V4-Flash is the exception, with only one of six combinations improving over both paradigms. This is consistent with the single-module results in \S~\ref{sec:marginal}: neither $G_2$ nor $G_3$ outperforms $G_1$ on this reasoning backbone, so the resulting combinations rarely outperform refinement alone.

The interaction terms reveal how these gains compose across backbones. On GPT-4o-mini, stacking usually outperforms the better individual paradigm, but all six interactions are sub-additive. This may reflect the weaker backbone's limited capacity to exploit generation and refinement jointly, yielding overlapping gains across stages. GPT-5.4 exhibits the opposite pattern: all six combinations outperform both individual paradigms, and four are super-additive.

\modulepara{Generation $\times$ Example Retrieval.} We further examine whether this backbone-dependent pattern extends to the input side through $G_2\times E_2$. The combination outperforms both individual paradigms on the three non-reasoning backbones, but not on DeepSeek-V4-Flash, where $G_2$ does not improve the baseline in isolation. $\mathcal{I}$ remains small across all four backbones, ranging from $-0.39$ to $+0.20$~pp, indicating near-additive composition for this module pairing overall.

\modulepara{Overall pattern.} Results from the two module pairings show that interaction effects differ systematically across backbones. \textit{\textbf{Module stacking improves accuracy on most backbones, but how its gains compose varies with backbone capability.}} Since results from two-paradigm combinations do not determine the performance of complete multi-module stacks, \S~\ref{sec:tradeoffs:stack} evaluates backbone-specific stacks and reports end-to-end accuracy and cost jointly.

\subsection{Stacking vs. Upgrading}
\label{sec:tradeoffs:stack}

\input{tables/table_stack}

\begin{figure}[!t]
\centering
\includegraphics[width=\columnwidth]{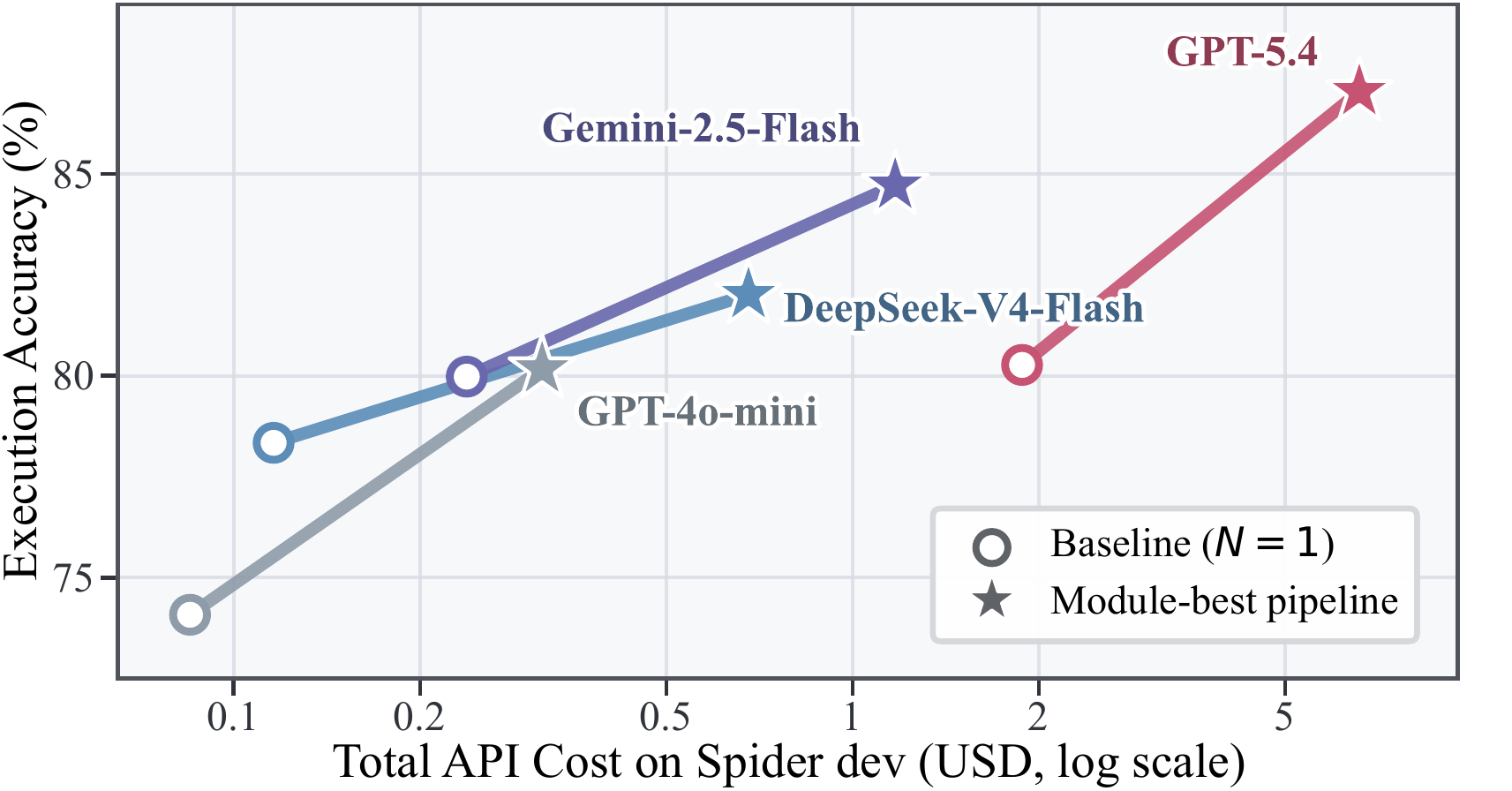}
\vspace{-6mm}
\caption{Cost--EX tradeoffs on Spider dev. Hollow circles denote $N{=}1$ baselines, filled stars denote the module-best pipelines selected on BIRD and applied unchanged to Spider. The horizontal axis reports total cost on a logarithmic scale.}
\label{fig:pareto}
\vspace{-5mm}
\end{figure}

\noindent\textbf{Stack construction rule.} We exclude Candidate Selection because its gains are bounded by candidate-pool diversity (\S~\ref{sec:marginal:dependent}) and do not justify sampling $N{>}1$ candidates from a single generator. For each backbone and each remaining module, we select the nontrivial paradigm with the lowest $\mathrm{CPP}$ whose $\mathrm{EX}$ gain is positive on that backbone and on a majority of backbones, retaining the trivial paradigm if none qualifies. This selects $S_2$ rather than the cheaper $S_1$ on GPT-4o-mini: although $S_1$ gains $+0.13$~pp on that backbone, its gain is negative on the other three. The resulting stacks in Table~\ref{tab:stack} largely follow the marginal analysis: all use $R_1$, while DeepSeek-V4-Flash retains $G_1$ because neither nontrivial generation paradigm improves its $\mathrm{EX}$. The only substantive departure from the marginal-$\mathrm{EX}$ ranking occurs in GPT-5.4's generation module, where the rule selects $G_2$ for its lower $\mathrm{CPP}$ despite $G_3$'s slightly higher $\mathrm{EX}$ overall.

\modulepara{BIRD tradeoff.} Table~\ref{tab:stack} summarizes the resulting stacks on BIRD. Stacking improves $\mathrm{EX}$ on every backbone, but by different margins: GPT-4o-mini, Gemini-2.5-Flash, and GPT-5.4 each gain more than $7$~pp, while DeepSeek-V4-Flash gains noticeably less, consistent with its weaker response to pipeline engineering observed in \S~\ref{sec:marginal}. \textit{\textbf{At moderate budgets, engineering a richer pipeline on a mid-tier backbone outperforms upgrading to a frontier backbone running lean: Gemini-2.5-Flash's stack Pareto-dominates GPT-5.4's lean baseline, achieving substantially higher $\mathrm{EX}$ at lower cost.}} This reframes choice as a backbone--pipeline tradeoff rather than a preference for the strongest model. At the high-budget end, GPT-5.4's stack attains the highest $\mathrm{EX}$, but its $2.29$~pp advantage over Gemini-2.5-Flash's stack comes at more than $5{\times}$ the total cost.

\modulepara{Spider generalization.} To test cross-benchmark transfer, we apply the BIRD-derived stacks unchanged to Spider. As shown in Figure~\ref{fig:pareto}, all four stacks improve $\mathrm{EX}$ over their respective lean baselines, with gains ranging from $+3.68$ to $+6.77$~pp. \textit{\textbf{The backbone--pipeline tradeoff persists on Spider: the DeepSeek-V4-Flash and Gemini-2.5-Flash stacks both Pareto-dominate GPT-5.4's lean baseline.}} The GPT-4o-mini stack nearly matches that baseline ($80.17\%$ vs.\ $80.27\%$) at \$0.315 rather than \$1.879, while the DeepSeek-V4-Flash and Gemini-2.5-Flash stacks surpass it, reaching $82.01\%$ and $84.72\%$ at lower costs of \$0.679 and \$1.172, respectively. GPT-5.4's stack achieves the highest $\mathrm{EX}$ of $87.04\%$, but its $2.32$~pp advantage over Gemini-2.5-Flash's stack requires $5.62\times$ the total cost overall.

\subsection{Cross-Module and -Model Conclusion}
\label{sec:tradeoffs:takeaways}

\textit{\textbf{Cross-module composition is backbone-dependent, and at moderate budgets the cost-efficient frontier runs through stacked mid-tier pipelines rather than lean configurations on frontier backbones.}} Marginal results remain useful for stack construction, but interactions shape how module gains combine. Applying the BIRD-derived stacks unchanged to Spider shows that the resulting backbone--pipeline tradeoff extends beyond BIRD. Table~\ref{tab:stack} reports a separate stack for each backbone; \S~\ref{sec:guideline} distills these results into a unified guideline and evaluates it on five diverse additional models.

%% file: tables/table_stack.tex
\begin{table}[!t]
\centering
\caption{Module-best pipelines on BIRD dev. Parentheses show $\Delta\mathrm{EX}$ (pp) over the $N{=}1$ baseline; API costs are in USD. Stacks follow the cost-aware rule in
\S~\ref{sec:tradeoffs:stack} and favor paradigms with low $\mathrm{CPP}$ and broadly positive gains across tested models.}
\label{tab:stack}
\vspace{-3mm}
\setlength{\tabcolsep}{3pt}

\resizebox{\columnwidth}{!}{%
\begin{tabular}{lccc}
\toprule
\textbf{Backbone}
& \textbf{Stack}
& \textbf{$\mathrm{EX}$ (\%)}$\uparrow$
& \textbf{Cost}$\downarrow$ \\
\midrule
GPT-4o-mini       & $E_2{+}S_2{+}G_2{+}R_1$ & 58.02\,(\textcolor{papergreen}{$+8.28$}) & 0.759 \\
DeepSeek-V4-Flash & $E_2{+}S_2{+}G_1{+}R_1$ & 61.15\,(\textcolor{papergreen}{$+4.82$}) & 1.584 \\
Gemini-2.5-Flash  & $E_2{+}S_2{+}G_2{+}R_1$ & 67.14\,(\textcolor{papergreen}{$+7.30$}) & 2.916 \\
GPT-5.4           & $E_2{+}S_2{+}G_2{+}R_1$ & 69.43\,(\textcolor{papergreen}{$+8.15$}) & 16.237 \\
\bottomrule
\end{tabular}%
}
\vspace{-4mm}
\end{table}

%% file: tex/guideline.tex
\section{Cost-Efficient Configuration Guideline}
\label{sec:guideline}
The per-backbone stacks in \S~\ref{sec:tradeoffs:stack} are backbone-specific selections derived from per-paradigm evaluation and therefore cannot be assumed in advance when configuring a new system. The practical question is instead whether the recurring patterns across these stacks can be compressed into a small set of configurations that requires only coarse knowledge of the backbone, rather than a new paradigm search for every model. This section distills the preceding single-module and stacking results into such a prescriptive guideline, examines the empirical patterns behind its recommendations, and directly evaluates its transfer to five additional validation backbones spanning diverse capability levels and reasoning behaviors.

\subsection{From Stacks to a Tiered Guideline}
\label{sec:guideline:extraction}

We distill the single-module and stacking results in our study into three clear incremental tiers for practical use. \textbf{Tier 1} is the bare backbone running a single-pass baseline. \textbf{Tier 2} adds $R_1$ (Execution-Feedback Refinement), the only paradigm that improves all four primary backbones at consistently low cost. \textbf{Tier 3} further adds $E_2$ (Masked Question Similarity) and $S_2$ (Pseudo-SQL Guided Linking), together with a backbone-conditional generation strategy: $G_2$ (CoT-Enhanced) for non-reasoning backbones and $G_1$ (trivial) for reasoning-capable ones, following the split identified in \S~\ref{sec:marginal:dependent}.

\input{tables/table_validation}

These choices are not obtained by simply taking the highest-accuracy paradigm in each module. They preserve the components that recur across the cost-efficient stacks in \S~\ref{sec:tradeoffs:stack}. All four stacks contain $E_2$, $S_2$, and $R_1$, whereas Generation Strategy is the main component that changes with backbone reasoning behavior: the three non-reasoning stacks use $G_2$, while DeepSeek-V4-Flash retains $G_1$. This recurring structure in turn motivates a fixed tier definition with the generation choice conditioned on the backbone. \textit{\textbf{The resulting guideline recommends a tier from the backbone profile and accuracy--cost preference, without requiring per-paradigm search for a new backbone.}} Figure~\ref{fig:guidelines} summarizes the resulting tier construction and backbone-conditioned selection rule.

\subsection{Patterns Behind the Guideline}
\label{sec:guideline:reading}

\textit{\textbf{Pipeline engineering is most cost-effective on small and mid-tier non-reasoning backbones.}} The Tier~1 to Tier~3 improvement is largest on the smallest backbone in our study, while a mid-tier non-reasoning backbone at Tier~3 costs a fraction of a frontier backbone at Tier~1 (\S~\ref{sec:tradeoffs:stack}). For example, Gemini-2.5-Flash at Tier~3 reaches higher EX than GPT-5.4 at Tier~1 while costing substantially less in our setting. The guideline therefore favors Tier~3 for these backbones, while using Tier~2 as the cost-efficient default for reasoning backbones and reserving Tier~3 on frontier backbones for accuracy-first settings; \S~\ref{sec:guideline:validation} tests this pattern on five backbones.

\input{tables/table_guideline}

\subsection{Validation on Additional Backbones}
\label{sec:guideline:validation}

To test whether the guideline transfers beyond the four primary backbones from which its tier structure was derived, we directly apply the three prescribed tiers to five additional validation backbones spanning different capability levels and reasoning behaviors, without conducting any per-paradigm search or adapting the tier composition to individual models. \textit{\textbf{All five validation backbones consistently improve from Tier~1 to both Tier~2 and Tier~3.}} As shown in Table~\ref{tab:guideline-validation}, Tier~2 gains range from +1.24 to +2.61 pp, while Tier~3 gains range from +3.98 to +7.63 pp. The improvement is observed across capability levels and both non-reasoning and reasoning settings, indicating that the fixed tier compositions continue to transfer beyond the four primary backbones used for stack construction. The corresponding costs also follow the intended tier ordering, with Tier~2 adding limited overhead and Tier~3 trading a larger cost increase for higher $\mathrm{EX}$ across five validation backbones.

%% file: tables/table_validation.tex
\begin{table}[!t]
\centering
\caption{Guideline validation on additional backbones. $\mathrm{EX}$ (\%) and total API cost in USD are measured on BIRD; parenthesized values report $\Delta\mathrm{EX}$ in percentage points over Tier~1. T1--T3 denote three progressively richer configurations, ranging from the lean baseline to the full cost-aware pipeline.}
\label{tab:guideline-validation}
\vspace{-2mm}
\setlength{\tabcolsep}{3pt}
\renewcommand{\arraystretch}{1.15}

\resizebox{1.0\columnwidth}{!}{%
\begin{tabular}{lcccccc}
\toprule
& \multicolumn{3}{c}{\textbf{$\mathrm{EX}$ (\%)}}
& \multicolumn{3}{c}{\textbf{Cost}} \\
\cmidrule(lr){2-4}
\cmidrule(lr){5-7}
\textbf{Backbone}
& \textbf{T1} & \textbf{T2} & \textbf{T3}
& \textbf{T1} & \textbf{T2} & \textbf{T3} \\
\midrule
Llama-3.1-8B-Instruct
& 37.61
& 40.22 {\footnotesize\textcolor{papergreen}{$(+2.61)$}}
& 44.07 {\footnotesize\textcolor{papergreen}{$(+6.46)$}}
& 0.05 & 0.06 & 0.09 \\
Qwen3-32B
& 50.59
& 52.80 {\footnotesize\textcolor{papergreen}{$(+2.21)$}}
& 55.48 {\footnotesize\textcolor{papergreen}{$(+4.89)$}}
& 0.43 & 0.52 & 0.93 \\
o3-mini
& 54.11
& 55.87 {\footnotesize\textcolor{papergreen}{$(+1.76)$}}
& 59.97 {\footnotesize\textcolor{papergreen}{$(+5.87)$}}
& 5.67 & 6.40 & 12.67 \\
Qwen3-235B
& 57.82
& 59.06 {\footnotesize\textcolor{papergreen}{$(+1.24)$}}
& 61.80 {\footnotesize\textcolor{papergreen}{$(+3.98)$}}
& 4.55 & 5.76 & 11.36 \\
Claude Sonnet 4.5
& 62.71
& 65.25 {\footnotesize\textcolor{papergreen}{$(+2.54)$}}
& 70.34 {\footnotesize\textcolor{papergreen}{$(+7.63)$}}
& 8.62 & 9.65 & 18.94 \\
\bottomrule
\end{tabular}%
}
\vspace{-2mm}
\end{table}

%% file: tables/table_guideline.tex
\definecolor{tierblue}{HTML}{DCE2F1}

\begin{figure}[t]
\centering
\begingroup
\small
\setlength{\fboxsep}{4pt}

\newcommand{\tierline}[1]{%
  \noindent
  $\hookrightarrow$\hspace{4pt}%
  {\setlength{\fboxsep}{2pt}%
  \colorbox{tierblue}{\strut\textbf{#1}}}%
  \par
}

\fcolorbox{black!35}{black!2}{%
\begin{minipage}{0.91\columnwidth}
\setlength{\parindent}{0pt}
\setlength{\parskip}{0pt}
\renewcommand{\baselinestretch}{1.06}\selectfont

\vspace*{2pt}

\textbf{[Tier Construction]}

\vspace{2pt}

\textbf{Bare-backbone baseline.}
Use the single-pass baseline with all modules at their trivial settings.

\vspace{1pt}
\tierline{T1: Bare backbone ($N{=}1$).}

\vspace{2pt}

\textbf{Universal refinement rule.}
$R_1$ is the only paradigm that improves all tested backbones at
consistently low cost.

\vspace{1pt}
\tierline{T2: $\mathrm{T1}+R_1$.}

\vspace{2pt}

\textbf{Backbone-conditioned stacking rule.}
$E_2+S_2$ persist across backbone stacks, while generation follows
the reasoning split:
\vspace{-2pt}
\[
G^{*}=
\begin{cases}
G_2, & \text{non-reasoning backbone},\\
G_1, & \text{reasoning backbone}.
\end{cases}
\]
\vspace{1pt}

\tierline{T3: $\mathrm{T2}+E_2+S_2+G^{*}$.}

\vspace{2pt}
\noindent\rule{\linewidth}{0.4pt}\par
\vspace{-2pt}

\textbf{[Candidate Selection]}

\vspace{1.5pt}

\textbf{Cost-efficient setting.}
Candidate Selection faces a fixed-pool recall ceiling and
$N{>}1$ sampling overhead, so it is excluded from the
cost-efficient tiers.

\vspace{1pt}

\textbf{Accuracy-first option.}
When cost is secondary, Candidate Selection may be added for further
accuracy, subject to the candidate-pool recall ceiling.

\vspace{1pt}
\noindent\rule{\linewidth}{0.4pt}\par
\vspace{-2pt}

\textbf{[Tier Selection]}

\vspace{1.5pt}

\renewcommand{\arraystretch}{1.05}
\begin{tabular}{@{}p{0.27\linewidth}p{0.67\linewidth}@{}}
\textbf{Reasoning} &
T2 is the cost-efficient default; consider T3 when extra accuracy
justifies the cost. \\[1.2pt]

\textbf{Small / mid-tier}\newline
\textbf{non-reasoning} &
Prefer T3. \\[1.2pt]

\textbf{Frontier} &
Use T3 when maximum EX is prioritized; otherwise consider a cheaper
backbone with T3.
\end{tabular}

\vspace*{2pt}

\end{minipage}
}
\vspace{-1mm}
\caption{Evidence-backed cost-efficient configuration guideline distilled from the single-module and stacking analyses. The three tiers progressively incorporate paradigms supported by robust accuracy--cost evidence, while Candidate Selection is treated separately due to its sampling overhead and fixed-pool recall ceiling. The recommended tier further depends on backbone capability and reasoning behavior.}
\label{fig:guidelines}
\vspace{-2mm}
\endgroup
\end{figure}

%% file: tex/related.tex
\section{Related Work}
\label{sec:related}

\noindent\textbf{Empirical Analysis of Text-to-SQL.} Several studies analyze specific components or configurations of Text-to-SQL systems. \citet{li2024dawn} introduce NL2SQL360, a multi-angle evaluation framework that compares NL2SQL methods across data domains, SQL characteristics, and evaluation settings, and further explores the selection of solutions for different application needs. At the component level, \citet{maamari2024death} revisit explicit schema linking and show that stronger LLMs can often operate directly over the full schema when it fits within the context window, avoiding errors caused by filtering necessary elements. In contrast, \citet{yuan2025knapsack} improve schema linking by formulating the selection of relevant schema elements as a constrained knapsack optimization. For generation, \citet{tai2023exploring} also show that iterative least-to-most prompting can be unnecessary and that overly detailed reasoning may propagate errors.

Recent systems increasingly combine multiple such mechanisms into elaborate inference-time pipelines. CHASE-SQL~\citep{pourreza2025chase} combines multi-path candidate generation with preference-based pairwise selection; DeepEye-SQL~\citep{li2026deepeye} organizes linking, diverse generation, verification, and candidate selection through a workflow; and Agentar-Scale-SQL~\citep{wang2025agentarscalesql} orchestrates internal reasoning, iterative refinement, and parallel generation and selection for test-time scaling. These systems demonstrate the benefit of increasingly sophisticated pipeline orchestration, but still primarily optimize integrated system designs and evaluate components within their own pipelines. Our study complements these efforts by isolating each module's marginal accuracy--cost contribution under one controlled implementation across multiple backbones; the schema-linking and CoT trends further extend prior observations to this cost-aware setting.

\modulepara{Cost-Aware LLM Systems.} Cost-aware LLM deployment has been studied at three complementary layers. At the \emph{model-selection} layer, FrugalGPT~\citep{chen2024frugalgpt}, SATER~\citep{shen2025sater}, and RouteLLM~\citep{ong2025routellm} use routing or cascading to navigate tradeoffs between performance and cost. At the \emph{inference-resource} layer, LLMLingua~\citep{jiang2023llmlingua} and Sketch-of-Thought~\citep{aytes2025sketch} reduce prompt or reasoning token usage, while recent work studies compute-optimal and budget-aware allocation of test-time reasoning resources~\citep{snell2025scaling,lin2026plan}. At the \emph{workflow} layer, CATP LLM~\citep{wu2025catp} explicitly balances task performance against tool-execution cost. Within Text-to-SQL, \citet{donder2025cheaper} propose N-rep, which reduces inference cost through multiple schema representations without CoT or self-consistency. These efforts optimize model choice, inference resources, or workflow execution; we instead directly ask which modules of the Text-to-SQL pipeline justify their added cost.

%% file: tex/conclusion.tex
\section{Conclusion}
\label{sec:conclusion}

We studied the ICL Text-to-SQL pipeline as a modular object in its own right, isolating the marginal contribution and cost of each design paradigm under a unified controlled framework across four backbones. Execution-feedback refinement is the only paradigm that improves all four primary backbones at consistently low cost, making it a safe first addition; most others pay off only under backbone-dependent conditions. Token accounting suggests that input demand is closely tied to pipeline structure, whereas output demand is sensitive to backbone generation behavior. Building on this analysis, we found that a fixed budget is often better spent engineering a richer pipeline over a mid-tier backbone than upgrading to a frontier model with a lean pipeline. The BIRD-derived stacks also transfer to Spider, where all four improve accuracy and the backbone--pipeline tradeoff persists. We distill these observations into a backbone-dependent, tiered guideline that transfers to five additional validation backbones, offering practitioners a more practical, cost-aware starting point for configuring ICL Text-to-SQL systems than aggregate accuracy alone in practical settings.

%% file: tex/ethical_considerations.tex
\section{Ethical Considerations}
\label{sec:ethical-considerations}

This work evaluates text-to-SQL methods on the publicly available BIRD benchmark~\citep{li2023BIRD}. Nevertheless, deployed text-to-SQL systems, especially in real world database applications, can expose sensitive database contents or execute harmful queries when model outputs are incorrect or maliciously prompted. Practical deployments should therefore enforce least-privilege access, use read-only execution whenever possible, validate generated SQL, sandbox query execution, and maintain proper human oversight and audit logs.

%% file: tex/appendix/full_taxonomy.tex
\newcommand{\NA}{\multicolumn{1}{c}{\textendash}}
\newcommand{\dexci}[2]{#1\hspace{0.9em}[#2]}

\twocolumn[{%
\begin{minipage}{\textwidth}
\centering
\captionof{table}{Per-backbone EX (\%) on BIRD dev (1{,}534 examples). $\Delta\mathrm{EX}$ [95\% CI] is the percentage-point gain over the matched-budget baseline ($N{=}5$ for Candidate Selection; $N{=}1$ otherwise), and CPP is incremental USD/pp. ``\textendash'' denotes undefined CPP, ``$^{*}$'' Pareto domination, and the \bestkey~/~\secondkey~keys mark the best/second best results.}
\label{tab:main-accuracy}
\vspace{-4mm}
\scriptsize
\setlength{\tabcolsep}{2.0pt}
\renewcommand{\arraystretch}{1.0}

\begin{tabular*}{\linewidth}{
@{\hspace{8pt}\extracolsep{\fill}}
c
c@{\hspace{3pt}}c@{\hspace{4pt}}c
c@{\hspace{3pt}}c@{\hspace{4pt}}c
c@{\hspace{3pt}}c@{\hspace{4pt}}c
c@{\hspace{3pt}}c@{\hspace{4pt}}c
@{\hspace{8pt}}
}

\toprule
\multirow{2}{*}{\textbf{ID}}
  & \multicolumn{3}{c}{\textbf{GPT-4o-mini}}
  & \multicolumn{3}{c}{\textbf{DeepSeek-V4-Flash}}
  & \multicolumn{3}{c}{\textbf{Gemini-2.5-Flash}}
  & \multicolumn{3}{c}{\textbf{GPT-5.4}} \\

\cmidrule(lr){2-4}
\cmidrule(lr){5-7}
\cmidrule(lr){8-10}
\cmidrule(lr){11-13}

& {$\mathbf{EX}\boldsymbol{\uparrow}$}
& {$\boldsymbol{\Delta}\mathbf{EX}\boldsymbol{\uparrow}$
   \hspace{0.9em}\textbf{[95\% CI]}}
& {$\mathbf{CPP}\boldsymbol{\downarrow}$}

& {$\mathbf{EX}\boldsymbol{\uparrow}$}
& {$\boldsymbol{\Delta}\mathbf{EX}\boldsymbol{\uparrow}$
   \hspace{0.9em}\textbf{[95\% CI]}}
& {$\mathbf{CPP}\boldsymbol{\downarrow}$}

& {$\mathbf{EX}\boldsymbol{\uparrow}$}
& {$\boldsymbol{\Delta}\mathbf{EX}\boldsymbol{\uparrow}$
   \hspace{0.9em}\textbf{[95\% CI]}}
& {$\mathbf{CPP}\boldsymbol{\downarrow}$}

& {$\mathbf{EX}\boldsymbol{\uparrow}$}
& {$\boldsymbol{\Delta}\mathbf{EX}\boldsymbol{\uparrow}$
   \hspace{0.9em}\textbf{[95\% CI]}}
& {$\mathbf{CPP}\boldsymbol{\downarrow}$} \\

\multicolumn{13}{l}
{\textbf{\textit{Example Retrieval}}} \\
\midrule
\addlinespace[2pt]

E$_1$
& \secondval{50.07} & \dexci{+0.33}{-0.01, +0.67} & \secondval{0.207}
& 56.06 & \dexci{-0.26}{-0.57, +0.05} & \NA
& \secondval{60.43} & \dexci{+0.59}{+0.06, +1.12} & \bestval{0.401}
& 62.58 & \dexci{+1.30}{+0.63, +1.97} & \secondval{0.995} \\

E$_2$
& \bestval{50.65} & \dexci{+0.91}{+0.26, +1.56} & \bestval{0.144}
& \secondval{56.84} & \dexci{+0.52}{+0.01, +1.03} & \bestval{0.326}
& \bestval{61.28} & \dexci{+1.43}{+0.74, +2.13} & \secondval{0.417}
& \bestval{63.69} & \dexci{+2.41}{+1.42, +3.40} & \bestval{0.957} \\

E$_3$
& 49.80 & \dexci{+0.07}{-0.16, +0.29} & 7.259
& \bestval{57.11} & \dexci{+0.78}{+0.06, +1.50} & \secondval{0.660}
& 60.10 & \dexci{+0.26}{-0.18, +0.70} & 5.317
& \secondval{63.49} & \dexci{+2.22}{+1.28, +3.15} & 3.417 \\

\multicolumn{13}{l}
{\textbf{\textit{Schema Linking}}} \\
\midrule
\addlinespace[2pt]

S$_1$
& 49.87 & \dexci{+0.13}{-0.13, +0.39} & \NA
& 55.93 & \dexci{-0.39}{-0.75, -0.03} & \NA
& 59.13 & \dexci{-0.72}{-1.27, -0.16} & \NA
& 60.63 & \dexci{-0.65}{-1.19, -0.11} & \NA \\

S$_2$
& \secondval{52.15} & \dexci{+2.41}{+1.49, +3.33} & \bestval{0.075}
& \bestval{57.04} & \dexci{+0.72}{+0.08, +1.36} & \bestval{0.491}
& \secondval{61.15} & \dexci{+1.30}{+0.63, +1.98} & \bestval{0.575}
& \bestval{62.39} & \dexci{+1.11}{+0.45, +1.77} & \bestval{3.293} \\

S$_{3a}$
& 51.11 & \dexci{+1.37}{+0.69, +2.05} & 0.109
& 56.19 & \dexci{-0.13}{-0.39, +0.13} & \NA
& 60.89 & \dexci{+1.04}{+0.42, +1.67} & \secondval{0.599}
& 60.82 & \dexci{-0.46}{-0.95, +0.03} & \NA \\

S$_{3b}$
& 51.83 & \dexci{+2.09}{+1.23, +2.95} & \secondval{0.090}
& 55.87 & \dexci{-0.46}{-0.88, -0.04} & \NA
& 60.76 & \dexci{+0.91}{+0.32, +1.51} & 0.847
& 59.97 & \dexci{-1.30}{-1.98, -0.63} & \NA \\

S$_4$
& \bestval{52.93} & \dexci{+3.19}{+2.11, +4.27} & 0.271
& \secondval{56.52} & \dexci{+0.20}{-0.09, +0.49} & \secondval{4.262}
& \bestval{61.34} & \dexci{+1.50}{+0.79, +2.21} & 1.224
& \secondval{62.06} & \dexci{+0.78}{+0.21, +1.35} & \secondval{20.870} \\

\multicolumn{13}{l}
{\textbf{\textit{SQL Generation}}} \\
\midrule
\addlinespace[2pt]

G$_1$
& 49.74 & 0.00 & \NA
& \bestval{56.32} & 0.00 & \NA
& 59.84 & 0.00 & \NA
& 61.28 & 0.00 & \NA \\

G$_2$
& \bestval{53.98} & \dexci{+4.24}{+2.96, +5.52} & \bestval{0.026}
& 55.93 & \dexci{-0.39}{-0.75, -0.03} & \NA
& \bestval{62.71} & \dexci{+2.87}{+1.80, +3.94} & \bestval{0.170}
& \secondval{62.84} & \dexci{+1.56}{+0.80, +2.32} & \bestval{1.723} \\

G$_3$
& \secondval{51.24} & \dexci{+1.50}{+0.77, +2.23} & \secondval{0.093}
& \secondval{56.13} & \dexci{-0.20}{-0.49, +0.09} & \NA
& \secondval{60.76} & \dexci{+0.91}{+0.32, +1.51} & \secondval{0.976}
& \bestval{63.10} & \dexci{+1.83}{+1.00, +2.65} & \secondval{2.285} \\

\multicolumn{13}{l}
{\textbf{\textit{Candidate Selection}}} \\
\midrule
\addlinespace[2pt]

C$_1$
& \secondval{51.11} & \dexci{+1.43}{+0.74, +2.13} & \bestval{$^{*}$}
& 57.30 & \dexci{+0.85}{+0.19, +1.51} & \bestval{$^{*}$}
& 60.89 & \dexci{+0.85}{+0.27, +1.43} & \bestval{$^{*}$}
& 62.71 & \dexci{+1.30}{+0.63, +1.97} & \bestval{$^{*}$} \\

C$_{2a}$
& 50.52 & \dexci{+0.85}{+0.36, +1.34} & \secondval{0.771}
& \bestval{59.00} & \dexci{+2.54}{+1.47, +3.61} & \secondval{0.333}
& \secondval{61.60} & \dexci{+1.56}{+0.80, +2.32} & \secondval{1.806}
& \secondval{63.82} & \dexci{+2.41}{+1.42, +3.40} & \secondval{7.249} \\

C$_{2b}$
& \bestval{51.17} & \dexci{+1.50}{+0.79, +2.21} & 1.389
& \secondval{58.54} & \dexci{+2.09}{+1.14, +3.04} & 0.953
& \bestval{61.93} & \dexci{+1.89}{+1.04, +2.74} & 2.858
& \bestval{63.89} & \dexci{+2.48}{+1.48, +3.48} & 15.975 \\

\multicolumn{13}{l}
{\textbf{\textit{Refinement}}} \\
\midrule
\addlinespace[2pt]

R$_1$
& \bestval{54.63} & \dexci{+4.89}{+3.53, +6.25} & \bestval{0.016}
& \secondval{58.15} & \dexci{+1.83}{+0.99, +2.67} & \bestval{0.040}
& \secondval{62.32} & \dexci{+2.48}{+1.48, +3.48} & \bestval{0.026}
& \secondval{63.30} & \dexci{+2.02}{+1.15, +2.89} & \bestval{0.297} \\

R$_2$
& 51.83 & \dexci{+2.09}{+1.23, +2.95} & 0.305
& 56.58 & \dexci{+0.26}{-0.05, +0.57} & 2.708
& 60.95 & \dexci{+1.11}{+0.45, +1.77} & 1.964
& 62.06 & \dexci{+0.78}{+0.21, +1.35} & 18.680 \\

R$_3$
& \secondval{53.26} & \dexci{+3.52}{+2.38, +4.66} & \secondval{0.276}
& \bestval{59.97} & \dexci{+3.65}{+2.47, +4.83} & \secondval{0.365}
& \bestval{63.04} & \dexci{+3.19}{+2.08, +4.30} & \secondval{1.016}
& \bestval{64.02} & \dexci{+2.74}{+1.68, +3.80} & \secondval{7.769} \\

\bottomrule
\end{tabular*}
\end{minipage}
\vspace{3mm}
}]

\section{Full Taxonomy Mapping}
\label{app:full-taxonomy}

Table~\ref{tab:text2sql_taxonomy} maps 24 ICL Text-to-SQL methods to paradigms across five pipeline modules. A checkmark indicates documented use of a paradigm, while a dash indicates that such use is not established in the original paper. For Direct LLM Linking and LLM-Based Preference Selection, sub-paradigm distinctions ($S_{3a}$ vs.\ $S_{3b}$ and $C_{2a}$ vs.\ $C_{2b}$) are recorded within cells rather than assigned separate columns.

\modulepara{Multi-paradigm cells.} A method may instantiate multiple paradigms within one module. This occurs mainly in Generation Strategy, where systems combine direct, CoT-enhanced, and decomposition-based generators, and in Schema Linking, where retrieval-based and LLM-based signals are combined before SQL generation.

\modulepara{Classification protocol.} We classify each method according to the primary system reported in the original paper, excluding ablation variants. When the paradigm is not stated explicitly, we infer it from the provided prompts or pseudocode; if neither provides sufficient evidence, the cell is marked as \na~in the final taxonomy table.

\begin{table*}[p]
\centering

\rotatebox[origin=c]{-90}{%
\begin{minipage}{0.96\textheight}
\centering
\small
\setlength{\tabcolsep}{7pt}
\renewcommand{\arraystretch}{1.55}
\setlength{\extrarowheight}{2.5pt}
\arrayrulecolor{black}
\setlength{\arrayrulewidth}{0.25pt}

\caption{Taxonomy of representative Text-to-SQL methods.}
\label{tab:text2sql_taxonomy}

\resizebox{\linewidth}{!}{%
\begin{tabular}{
  |C{3.4cm}|
  C{2.0cm}|
  C{1.9cm}|C{1.9cm}|C{1.9cm}|
  C{1.9cm}|C{1.9cm}|C{1.9cm}|C{1.9cm}|
  C{1.9cm}|C{1.9cm}|C{1.9cm}|
  C{1.9cm}|C{1.9cm}|
  C{1.9cm}|C{1.9cm}|C{1.9cm}|
}

\hline
\multirowcell{2}{\textbf{Method}} &
\multirowcell{2}{\textbf{Venue}} &
\multicolumn{3}{c|}{\textbf{Example Retrieval}} &
\multicolumn{4}{c|}{\textbf{Schema Linking}} &
\multicolumn{3}{c|}{\textbf{Generation Strategy}} &
\multicolumn{2}{c|}{\textbf{Candidate Selection}} &
\multicolumn{3}{c|}{\textbf{Refinement}} \\
\cline{3-17}
\noalign{\vskip 3pt}
&
&
\makecell{\textbf{Question}\\\textbf{Similarity}} &
\makecell{\textbf{Masked}\\\textbf{Question}\\\textbf{Similarity}} &
\makecell{\textbf{SQL}\\\textbf{Skeleton}\\\textbf{Similarity}} &
\makecell{\textbf{Retrieval-}\\\textbf{Based}\\\textbf{Linking}} &
\makecell{\textbf{Pseudo-SQL}\\\textbf{Guided}\\\textbf{Linking}} &
\makecell{\textbf{Direct LLM}\\\textbf{Linking}} &
\makecell{\textbf{Agentic}\\\textbf{Linking}} &
\makecell{\textbf{Direct}\\\textbf{Generation}} &
\makecell{\textbf{CoT-Enhanced}\\\textbf{Generation}} &
\makecell{\textbf{Decomposition-}\\\textbf{Based}\\\textbf{Generation}} &
\makecell{\textbf{Execution-}\\\textbf{Based}\\\textbf{Voting}} &
\makecell{\textbf{LLM-Based}\\\textbf{Preference}\\\textbf{Selection}} &
\makecell{\textbf{Execution-}\\\textbf{Feedback}\\\textbf{Refinement}} &
\makecell{\textbf{Self-}\\\textbf{Correction}} &
\makecell{\textbf{Agentic}\\\textbf{Refinement}} \\
\hline

AutoLink & AAAI'26
& \na & \na & \na
& \na & \na & \na & \cmark
& \na & \cmark & \cmark
& \cmark & \makecell{Pairwise\\Comparison}
& \cmark & \na & \na \\
\hline

DeepEye-SQL & SIGMOD'26
& \na & \cmark & \na
& \cmark & \cmark & \makecell{Single-shot\\Direct Linking} & \na
& \cmark & \cmark & \cmark
& \cmark & \makecell{Pairwise\\Comparison}
& \na & \na & \cmark \\
\hline

MAGIC & AAAI'25
& \na & \na & \na
& \na & \na & \na & \na
& \na & \na & \na
& \na & \na
& \na & \cmark & \na \\
\hline

Agentar-Scale-SQL & arXiv'25
& \na & \cmark & \na
& \na & \na & \na & \na
& \cmark & \cmark & \cmark
& \na & \makecell{Pairwise\\Comparison}
& \cmark & \cmark & \na \\
\hline

N-rep & arXiv'25
& \na & \cmark & \na
& \na & \na & \makecell{Single-shot\\Direct Linking} & \na
& \cmark & \na & \na
& \cmark & \makecell{Pairwise\\Comparison}
& \na & \na & \na \\
\hline

XiYan-SQL & arXiv'25
& \na & \cmark & \na
& \cmark & \na & \makecell{Single-shot\\Direct Linking} & \na
& \cmark & \na & \na
& \na & \makecell{Scoring\\and Ranking}
& \cmark & \na & \na \\
\hline

Gen-SQL & COLING'25
& \cmark & \na & \na
& \cmark & \cmark & \na & \na
& \cmark & \na & \na
& \na & \na
& \cmark & \na & \na \\
\hline

MCS-SQL & COLING'25
& \cmark & \cmark & \na
& \na & \na & \makecell{Two-stage\\Direct Linking} & \na
& \na & \cmark & \na
& \na & \makecell{Scoring\\and Ranking}
& \na & \na & \na \\
\hline

MAC-SQL & COLING'25
& \na & \na & \na
& \na & \na & \na & \cmark
& \na & \cmark & \cmark
& \na & \na
& \na & \na & \cmark \\
\hline

PET-SQL & DASFAA'25
& \na & \cmark & \na
& \na & \cmark & \na & \na
& \cmark & \na & \na
& \cmark & \na
& \na & \na & \na \\
\hline

LinkAlign & EMNLP'25
& \na & \na & \na
& \na & \na & \na & \cmark
& \na & \na & \na
& \na & \na
& \na & \na & \na \\
\hline

Alpha-SQL & ICML'25
& \na & \na & \na
& \cmark & \na & \makecell{Single-shot\\Direct Linking} & \na
& \na & \cmark & \cmark
& \cmark & \na
& \cmark & \na & \na \\
\hline

CHESS-SQL & ICML-W'25
& \na & \na & \na
& \na & \na & \makecell{Two-stage\\Direct Linking} & \na
& \na & \cmark & \na
& \na & \makecell{Scoring\\and Ranking}
& \cmark & \na & \na \\
\hline

CHASE-SQL & ICLR'25
& \na & \na & \na
& \na & \na & \makecell{Two-stage\\Direct Linking} & \na
& \na & \cmark & \cmark
& \cmark & \makecell{Pairwise\\Comparison}
& \na & \na & \cmark \\
\hline

ReFoRCE & ICLR-W'25
& \na & \na & \na
& \na & \na & \makecell{Single-shot\\Direct Linking} & \na
& \cmark & \na & \na
& \cmark & \na
& \na & \na & \cmark \\
\hline

OpenSearch-SQL & SIGMOD'25
& \na & \cmark & \na
& \cmark & \na & \makecell{Single-shot\\Direct Linking} & \na
& \na & \cmark & \na
& \cmark & \na
& \cmark & \na & \na \\
\hline

RSL-SQL & arXiv'24
& \cmark & \na & \na
& \na & \cmark & \makecell{Single-shot\\Direct Linking} & \na
& \cmark & \na & \cmark
& \na & \makecell{Pairwise\\Comparison}
& \cmark & \na & \na \\
\hline

DEA-SQL & ACL-F'24
& \cmark & \cmark & \na
& \na & \na & \makecell{Two-stage\\Direct Linking} & \na
& \cmark & \na & \na
& \na & \na
& \na & \cmark & \na \\
\hline

PURPLE & ICDE'24
& \na & \na & \cmark
& \cmark & \na & \na & \na
& \cmark & \na & \na
& \cmark & \na
& \na & \na & \na \\
\hline

SuperSQL & VLDB'24
& \cmark & \na & \cmark
& \cmark & \na & \na & \na
& \cmark & \na & \na
& \cmark & \na
& \na & \na & \na \\
\hline

DAIL-SQL & VLDB'24
& \cmark & \cmark & \na
& \na & \na & \na & \na
& \cmark & \na & \na
& \cmark & \na
& \na & \na & \na \\
\hline

C3-SQL & arXiv'23
& \na & \na & \na
& \na & \na & \makecell{Two-stage\\Direct Linking} & \na
& \cmark & \na & \na
& \cmark & \na
& \na & \na & \na \\
\hline

ACT-SQL & EMNLP-F'23
& \cmark & \na & \na
& \na & \na & \na & \na
& \na & \cmark & \na
& \na & \na
& \na & \na & \na \\
\hline

DIN-SQL & NeurIPS'23
& \na & \na & \na
& \na & \na & \makecell{Single-shot\\Direct Linking} & \na
& \na & \cmark & \cmark
& \na & \na
& \na & \cmark & \na \\
\hline

\end{tabular}%
}

\end{minipage}%
}

\end{table*}

%% file: tex/appendix/paradigm_implementation_details.tex
\section{Paradigm Implementation Details}
\label{app:impl-cards}

This section details the controlled implementations of the 17 paradigm-level configurations evaluated in the main paper. We organize them into pre-processing (Example Retrieval and Schema Linking; Figure~\ref{fig:impl-cards-1}), SQL generation (Generation Strategy; Figure~\ref{fig:impl-cards-2}), and post-processing (Candidate Selection and Refinement; Figure~\ref{fig:impl-cards-3}).

\modulepara{Box layout.}
Each module card specifies its shared inputs, backbone, and fallback behavior; maps paradigms to mechanisms and per-query LLM calls; notes auxiliary calls or behavior and their cost treatment; and identifies the trivial baseline used in the main marginal analysis for controlled comparison across paradigms..

\begin{figure*}[p]
\centering
\begin{paradigmbox}

\textbf{Example Retrieval.}\\[2pt]
\textbf{Shared setup.} All three paradigms retrieve from the BIRD training split using a frozen \texttt{bge-large-en-v1.5} encoder, with $\ell_2$-normalized embeddings ranked by cosine similarity. Examples drawn from the target database are excluded to prevent leakage, and exact-duplicate questions are removed. The generator receives up to $k{=}3$ demonstrations, each rendered as the original question, its optional evidence annotation, and the gold SQL; no demonstration schema is included. Embeddings are precomputed once and cached.

\smallskip
\begin{tabularx}{\linewidth}{@{}l X c@{}}
\toprule
Paradigm & Retrieval signal & Shots \\
\midrule
$E_1$ Question Similarity
& Raw question embedding.
& 3 \\
E$_2$ Masked Question Similarity
& Question with schema-specific entities masked by the backbone before
encoding; retrieval returns the original (unmasked) examples.
& 3 \\
E$_3$ SQL Skeleton Similarity
& Structural skeleton of a preliminary SQL draft, matched against
skeletons of training gold SQL.
& 3 \\
\bottomrule
\end{tabularx}
\smallskip

\textbf{Auxiliary LLM calls.} E$_2$ issues one backbone call per query to produce the masked question; E$_3$ issues one call to draft the preliminary SQL from which the skeleton is extracted. $E_1$ requires no auxiliary call. All auxiliary calls use the same backbone as the main generator, at temperature~0; on an empty or malformed response, the paradigm falls back to the unmasked question (E$_2$) or to direct generation (E$_3$). Skeleton extraction abstracts literals and non-keyword identifiers to placeholder tokens while preserving SQL keywords and operators.

\smallskip
\textbf{Trivial baseline.} Zero-shot generation with no retrieved
demonstrations.

\modulesep

\textbf{Schema Linking.}\\[2pt]
\textbf{Shared setup.} Each paradigm receives the question, the full schema with column descriptions, and the optional evidence
annotation, and returns a filtered schema for the generator; primary and foreign keys are always retained after filtering. Paradigms differ in who makes the linking decision. Malformed or empty responses fall back to the full schema, so no paradigm gains accuracy from its failure path. All auxiliary calls use the main backbone, so their cost is attributed to that backbone.

\smallskip
\begin{tabularx}{\linewidth}{@{}l X c@{}}
\toprule
Paradigm & Mechanism & Calls \\
\midrule
$S_1$ Retrieval-Based
& Embeds question keywords against per-column text, boosts columns whose cells match a keyword, and keeps the top-scoring tables and columns.
& 1 \\
S$_2$ Pseudo-SQL Guided
& Drafts a SQL query over the full schema, then keeps the referenced tables and columns.
& 1 \\
S$_{3a}$ Direct LLM, Single-shot
& Asks the backbone once to select the needed tables and columns, then validates the names against the schema.
& 1 \\
S$_{3b}$ Direct LLM, Two-stage
& Selects relevant tables first, then selects columns only within those
tables.
& 2 \\
S$_4$ Agentic
& A proposer selects a schema slice, a verifier judges its sufficiency, and an optional second proposer revises it.
& 1--3 \\
\bottomrule
\end{tabularx}
\smallskip

\textbf{Auxiliary LLM calls.} All linking calls run on the main backbone at temperature~0. $S_1$ uses a frozen
\texttt{bge-large-en-v1.5} encoder for its keyword--column matching.
S$_4$ issues a variable number of calls: a single proposer call when its proposal is empty or unparsable, two calls when the verifier accepts the slice, and a third only when the verifier reports a concrete deficiency.

\smallskip
\textbf{Trivial baseline.} The full schema is passed to the generator unfiltered.

\end{paradigmbox}
\caption{Controlled implementation of pre-processing modules: Example Retrieval and Schema Linking.}
\label{fig:impl-cards-1}
\end{figure*}

\begin{figure*}[p]
\centering
\begin{paradigmbox}

\textbf{Generation Strategy.}\\[2pt]
\textbf{Shared setup.} All three paradigms receive the question, the schema (linked or full), and the optional evidence, and run on the main backbone at temperature~0 with a single sampled candidate---one generation call per query. Each parses the final SQL string from its own output and passes only that string downstream; when no SQL can be parsed, the query counts as incorrect.

\smallskip
\begin{tabularx}{\linewidth}{@{}l X c@{}}
\toprule
Paradigm & Mechanism & Calls \\
\midrule
$G_1$ Direct
& The backbone writes a single SQL query directly, returned in a fenced code block.
& 1 \\
G$_2$ CoT-Enhanced
& The backbone emits a step-by-step reasoning trace and the final SQL in one structured response.
& 1 \\
G$_3$ Decomposition-Based
& The backbone decomposes the question into sub-questions and sub-queries and returns an assembled final SQL, all in one structured response.
& 1 \\
\bottomrule
\end{tabularx}
\smallskip

\textbf{Generation behavior.} All three paradigms use a single backbone call; they differ only in what that call is asked to emit. In G$_2$, reasoning and SQL are produced together and is therefore charged as additional output tokens rather than obtained from a separate planning call. In G$_3$, the number of sub-questions is chosen by the model, and the final SQL is read directly from its output rather than assembled by executing sub-queries; G$_3$ likewise adds no extra call. SQL is parsed from a structured field where available (G$_2$, G$_3$) or from the fenced block ($G_1$), with a shared fallback that extracts a bare \texttt{SELECT}/\texttt{WITH} statement.

\smallskip
\textbf{Trivial baseline.} Direct generation (G$_1$).

\end{paradigmbox}
\caption{Controlled implementation of the generation stage: Generation Strategy.}
\label{fig:impl-cards-2}
\end{figure*}

\begin{figure*}[p]
\centering
\begin{paradigmbox}

\textbf{Candidate Selection.}\\[2pt]
\textbf{Shared setup.} All selectors operate over the same pool of $N{=}5$ candidates, sampled from the direct-generation prompt on the main backbone at temperature~0.8---the same pool used by the $N{=}5$ reference baseline. Each selector returns one of the five candidates and never generates a new query.

\smallskip
\begin{tabularx}{\linewidth}{@{}l X c@{}}
\toprule
Paradigm & Mechanism & LLM calls \\
\midrule
$C_1$ Execution-Based Voting
& Executes all five candidates, groups them by result equivalence, and takes the earliest candidate in the largest group.
& 0 \\
C$_{2a}$ Scoring and Ranking
& Shows the model all five candidates with their execution outcomes and asks it to pick the best.
& 1 \\
C$_{2b}$ Pairwise Comparison
& Runs a single-elimination tournament, comparing the running winner against each remaining candidate.
& $N{-}1$ \\
\bottomrule
\end{tabularx}
\smallskip

\textbf{Selection behavior.} All selectors first execute the five candidates and run on the main backbone at temperature~0. They differ in cost: C$_1$ issues no LLM call (execution only), C$_{2a}$ a single call independent of $N$, and C$_{2b}$ $N{-}1$ calls (four at $N{=}5$), the dominant cost driver among the selectors.

\smallskip
\textbf{Trivial baseline.} Return the first candidate from the same $N{=}5$ candidate pool.

\modulesep

\textbf{Refinement.}\\[2pt]
\textbf{Shared setup.} Each paradigm receives the selected SQL, the question, the schema, and the optional evidence, and returns a final SQL that replaces the original for evaluation; if refinement yields no valid SQL, the original (or most recent valid revision) is kept. All refinement calls run on the main backbone at temperature~0.

\smallskip
\begin{tabularx}{\linewidth}{@{}l X c@{}}
\toprule
Paradigm & Mechanism & LLM calls \\
\midrule
$R_1$ Execution-Feedback
& Executes the SQL first, then asks the model to rewrite it only when execution errors or returns no rows.
& 0--1 \\
R$_2$ Self-Correction
& Asks the model to review the SQL against a generic error checklist, without executing it.
& 1 \\
R$_3$ Agentic
& A corrector revises the SQL using execution feedback, a verifier reviews the result, and the corrector may revise once more.
& 1--3 \\
\bottomrule
\end{tabularx}
\smallskip

\textbf{Refinement behavior.} The paradigms differ in when they fire, which is what separates their cost. R$_1$ is failure gated: it runs one diagnostic execution and issues a model call only on an error or empty result, so its average call count per query is below one---the reason it adds accuracy at near-zero cost. R$_2$ and R$_3$ instead fire on every query: R$_2$ issues exactly one review call with no execution feedback, while R$_3$ issues two to three calls across its corrector and verifier roles. Each paradigm performs at most one revision round (R$_3$ at most one additional corrector pass).

\smallskip
\textbf{Trivial baseline.} The selected SQL is returned unchanged, with no refinement call.

\end{paradigmbox}
\caption{Controlled implementation of post-processing modules: Candidate Selection and Refinement.}
\label{fig:impl-cards-3}
\end{figure*}

%% file: tex/appendix/pricing.tex
\section{API Pricing}
\label{app:pricing}

All USD costs reported in the main paper and this supplement are calculated using the API prices in Table~\ref{tab:pricing}, recorded on May~2, 2026. For each benchmark, the total configuration cost is the sum of input- and output-token charges over its full development split.

\begin{table}[H]
\centering
\vspace{-3mm}
\caption{API prices in USD per million tokens, recorded on May~2, 2026. For reasoning backbones, output charges include all billed reasoning tokens used in evaluation.}
\label{tab:pricing}
\vspace{-2mm}

\normalsize
\setlength{\tabcolsep}{3pt}
\renewcommand{\arraystretch}{0.9}

\begin{tabular*}{\columnwidth}{
@{\extracolsep{\fill}}lrr@{}
}
\toprule
\textbf{Model}
& \textbf{Input}
& \textbf{Output} \\
\midrule
GPT-4o-mini                & 0.15 & 0.60  \\
Gemini-2.5-Flash           & 0.30 & 2.50  \\
DeepSeek-V4-Flash          & 0.14 & 0.28  \\
GPT-5.4                    & 2.50 & 15.00 \\
Llama-3.1-8B-Instruct      & 0.02 & 0.04  \\
Qwen3-32B                  & 0.08 & 0.28  \\
o3-mini                    & 1.10 & 4.40  \\
Qwen3-235B-A22B (Thinking) & 0.23 & 2.30  \\
Claude Sonnet 4.5          & 3.00 & 15.00 \\
\bottomrule
\end{tabular*}
\end{table}

%% file: tex/appendix/main_results.tex
\section{Detailed Results with Confidence Intervals}
\label{app:results}

This section supplements the module-level results in Table~2 of the main paper. For every paradigm--backbone pair, Table~\ref{tab:main-accuracy} reports execution accuracy, its change from the matched-budget reference baseline with a corresponding 95\% confidence interval, and cost per $\mathrm{EX}$ point ($\mathrm{CPP}$). The main-paper table provides the corresponding API costs and mean per-question input/output-token counts.

\modulepara{Confidence intervals.} For each paradigm--backbone pair, we compare per-example correctness with a matched-budget reference over the 1{,}534 BIRD development examples. Two-sided 95\% confidence intervals for $\Delta\mathrm{EX}$ use the normal approximation to the paired difference based on discordant outcome counts, while significance is assessed by a two-sided McNemar's test at $\alpha{=}0.05$. Candidate Selection uses an $N{=}5$ reference, whereas all other paradigms use $N{=}1$, isolating the paradigm effect from additional candidate sampling. We report the intervals directly to preserve effect direction and magnitude rather than only binary significance decisions.